\documentclass[pdflatex,sn-mathphys-num]{sn-jnl}

\usepackage{tikz}
\usepackage{pgfplots}
\pgfplotsset{compat=1.18}
\usepackage{subcaption}
\usepackage[most]{tcolorbox}
\usepackage{graphicx}%
\usepackage{multirow}%
\usepackage{amsmath,amssymb,amsfonts}%
\usepackage{amsthm}%
\usepackage{mathrsfs}%
\usepackage[title]{appendix}%
\usepackage{xcolor}%
\usepackage{textcomp}%
\usepackage{manyfoot}%
\usepackage{booktabs}%
\usepackage{algorithm}%
\usepackage{algorithmicx}%
\usepackage{algpseudocode}%
\usepackage{listings}%
\usepackage{float}

\newcommand{\ignore}[1]{}
\newcommand{\rev}[1]{#1}

\theoremstyle{thmstyleone}%
\theoremstyle{thmstyletwo}%

\theoremstyle{thmstylethree}%

\begin{document}

\title[Article Title]{GenOM: Ontology Matching with Description Generation and Large Language Models}


\author*[1]{\fnm{Yiping} \sur{Song} }\email{yiping.song@postgrad.manchester.ac.uk}

\author[1]{\fnm{Jiaoyan} \sur{Chen}}\email{jiaoyan.chen@manchester.ac.uk}

\author[1]{\fnm{Renate A.} \sur{Schmidt}}\email{renate.schmidt@manchester.ac.uk}

\affil*[1]{
  \orgdiv{Department of Computer Science},
  \orgname{The University of Manchester},
  \orgaddress{
    \street{Oxford Road},
    \city{Manchester},
    \postcode{M13 9PL},
    \country{United Kingdom}
  }
}


\abstract{Ontology matching (OM) plays an essential role in enabling semantic interoperability and integration across heterogeneous knowledge sources, particularly in biomedical domains which contain numerous complex concepts related to diseases and pharmaceuticals. This paper introduces \textbf{GenOM}, a large language model (LLM)-based ontology alignment framework, which enriches  semantic representations of ontology concepts via generating textual definitions, retrieving alignment candidates with an embedding model, and incorporating exact lexical matching tools to improve precision. Extensive experiments conducted on the OAEI Bio-ML track demonstrate that GenOM can often achieve competitive performance, surpassing many baselines including traditional OM systems and recent LLM-based methods. Further ablation studies confirm the effectiveness of semantic enrichment, highlighting the framework’s robustness and adaptability. Beyond the matching framework itself, this paper introduces a set of criteria for evaluating the quality of concept definitions that are generated, providing a more systematic basis for analysing LLM-generated descriptions.}




\maketitle
\begin{center}
\small
This paper has been accepted for publication in \textit{World Wide Web} (Springer), 2026.\\
This version includes revisions based on peer review.
\end{center}
\vspace{0.5em}

\section{Introduction}

In recent years, the rapid growth of domain-specific ontologies has intensified the demand for \textit{semantic interoperability} and \textit{knowledge integration} across heterogeneous information systems~\cite{staab2013handbook}. 
In particular, the biomedical domain~\cite{lambrix2009information} has witnessed the proliferation of large and complex ontologies, such as those used to model diseases, anatomy, and clinical procedures, which further amplifies this challenge.
While ontologies serve as structured and machine-interpretable representations of conceptual knowledge, they are often engineered independently by different research communities or organisations. 
Each ontology typically reflects particular domain-specific objectives, modelling assumptions, and terminological conventions, which collectively result in substantial structural and semantic heterogeneity. 
This diversity, though valuable for capturing local perspectives, poses significant challenges to semantic interoperability and the effective integration of knowledge across heterogeneous information systems. 
Consequently, bridging conceptual discrepancies among ontologies has become a central challenge for achieving seamless knowledge exchange and reuse in data-intensive environments. Against this backdrop, this paper aims to develop a framework that leverages the reasoning and generation capabilities of large language models (LLMs) to support ontology matching across heterogeneous ontologies.

\rev{Ontology matching (OM) is widely recognized as a fundamental technique for enabling semantic interoperability across heterogeneous ontologies~\cite{6104044,euzenat2013d}.} 
Its primary objective is to uncover semantic correspondences between entities that originate from different ontologies, thereby enabling the consistent interpretation and exchange of knowledge. 
Through establishing formal relations such as \textit{equivalence}, where two entities express the same or highly similar meanings, and \textit{subsumption}, where one entity denotes a more general or more specific concept, OM serves as a cornerstone for semantic interoperability in distributed information environments.

Nevertheless, achieving accurate alignments remains highly challenging due to the intrinsic heterogeneity of ontologies. 
This heterogeneity is manifested in several forms. 
First, \textit{terminological variation} arises when identical concepts are described using distinct lexical labels or synonyms. 
Second, \textit{structural variation} reflects the diversity of ontology design choices, ranging from deep hierarchical taxonomies to flat enumerative structures. 
Third, \textit{granularity variation} occurs when similar knowledge is modelled at differing levels of abstraction. 
Together, these inconsistencies amplify both the cognitive and computational complexity of the matching task. 
The challenge becomes even more pronounced as ontologies continue to scale. 
For example, SNOMED CT~\cite{snomedct2024}, one of the largest biomedical terminologies, comprises hundreds of thousands of concepts, which has led some studies to further decompose it into smaller, task-specific subontologies for improved tractability~\cite{delpinto2024ips,del-pinto_extracting_2022}, yet purely manual alignment remains impractical at this scale, further motivating the need for scalable automated solutions.

Over the past decade, a range of automated systems, such as LogMap~\cite{jimenez2011logmap} and AgreementMakerLight (AML)~\cite{faria2013agreementmakerlight}, have been proposed to address this challenge. 
These approaches integrate lexical similarity, structural heuristics, and background knowledge resources to identify candidate mappings. 
While effective in well-structured and lexically rich scenarios, they often fail to capture the deeper semantics encoded in complex concept descriptions. 
\rev{Recent progress in LLMs has opened new possibilities for addressing this limitation~\cite{he2023exploring}.} 
By virtue of their powerful contextual reasoning and linguistic generalisation abilities, LLMs offer an unprecedented opportunity to represent and compare ontological knowledge at a semantic level~\cite{brown2020language}. 
Several contemporary OM systems have begun to incorporate LLMs at different stages of the alignment pipeline, including semantic embedding and direct alignment prediction.
For instance, LLM4OM~\cite{giglou2024llms4om} integrates ChatGPT with OpenAI embeddings to support pairwise concept matching, while Olala~\cite{hertling2023olala} leverages LLM-based semantic reasoning to assess concept equivalence. A more detailed review is given in Section~\ref{sec2}. 
Despite these advances, current LLM-based approaches remain constrained by practical and methodological limitations. 
Some rely on extremely large models (e.g., the 70B-parameter LLaMA-2 employed in Olala~\cite{hertling2023olala}), which incur considerable computational costs and hinder scalability, whereas others~\cite{Norouzi2023ConversationalOA} struggle to maintain robustness when confronted with complex or large-scale alignment scenarios.

To address these limitations, we develop \textbf{GenOM}, an ontology matching framework that leverages LLMs for semantic enrichment, candidate generation, and mapping judgement.\footnote{A preliminary version of this idea, focusing on LLM-based definition generation, was presented at the OM Workshop @ ISWC 2025~\cite{song2025genom}.} GenOM begins by extracting lexical and structural information from both source and target ontologies—including labels, synonyms, parent relations, and equivalence axioms—and employs an LLM-driven description generation module to produce semantically enriched textual definitions for each concept. These enriched descriptions are then embedded into a vector space to support efficient retrieval of semantically similar candidates, while an LLM-based mapping judgement component and a lightweight exact-matching module jointly refine the alignments.

In this work, we instantiate GenOM using a diverse suite of LLMs, which vary in architectural design and parameter scale. This enables a systematic analysis of how model capacity influences definition quality, semantic representation, and overall alignment performance.

We conduct a comprehensive experimental study across five OM tasks involving biomedical ontologies of SNOMED-CT, NCIT, FMA, ORDO, DOID, and OMIM (see Section \ref{sec:dataset} for more details). 
Our evaluation examines GenOM from four perspectives: 
(1) its overall alignment performance is compared with strong existing systems; 
(2) the quality of LLM-generated definitions is assessed using lexical metrics and an LLM-as-a-judge semantic evaluation; 
(3) performance differences across LLMs of varying scales and architectures are analysed; 
(4) ablation studies are conducted to isolate the contributions of semantic enrichment, candidate generation, and mapping judgement. 
These experiments collectively demonstrate the robustness of GenOM and the importance of LLM-driven semantic enrichment in ontology matching. An open-source implementation of the proposed approach is available.\footnote{Code repository: \url{https://github.com/AlanS87/genom-pipeline}}
\\
\\
\textbf{Contributions:} 
This work makes the following key contributions:

\begin{itemize}
    \item \textbf{GenOM: an LLM-enhanced ontology matching framework.}
    We introduce GenOM, a new ontology matching framework that integrates LLM-driven semantic enrichment with embedding-based candidate generation and LLM-based mapping judgement. \rev{Unlike previous approaches that rely primarily on existing ontology annotations, GenOM exploits the conceptual knowledge implicitly stored in large language models to generate informative definitions for ontology concepts. This capability is particularly beneficial for concepts whose ontology descriptions are sparse or incomplete, for example when only a label is available. By enriching such concepts with LLM-generated definitions, GenOM provides additional semantic signals that substantially improve the ability of the downstream alignment model to distinguish correct mappings from closely related alternatives, as demonstrated in our experiments.}

    \item \textbf{A novel evaluation methodology for assessing the quality of LLM-generated concept definitions.}
    \rev{Since the generated definitions directly influence candidate retrieval and equivalence judgement, it is important to assess their semantic correctness and discriminative ability.} We propose a dedicated evaluation scheme for concept definitions in OM tasks, combining lexical metrics, alignment-based measures, and an LLM-as-a-judge semantic assessment to quantify the correctness and discriminativeness of generated definitions.

    \item \textbf{A comparative analysis of LLMs of different scales for ontology matching.}
    We systematically compare LLMs over varying parameter sizes and architectures, revealing how model capacity influences definition quality, threshold behaviour, and final alignment performance, and highlighting their potential for future ontology matching research.
\end{itemize}

The remainder of this paper is organised as follows.
Section~\ref{sec2} reviews related work on ontology matching.
Section~\ref{sec3} introduces the GenOM framework in detail, including its overall methodology and the design of each core component.
We report the performance of GenOM on multiple biomedical ontology matching tasks, assess it from multiple perspectives, and compare it against several strong baseline systems in Section~\ref{sec4}.
Section~\ref{sec5} presents an in-depth analysis of the quality of LLM-generated definitions, examining their correctness and discriminative capacity.
Section~\ref{sec6} investigates the impact of different LLMs, analysing how model scale and architecture influence alignment behaviour and performance.
Section~\ref{sec7} provides an ablation study to analyse the impact of LLM-generated definitions on different stages of the alignment process.
Finally, Section~\ref{sec8} concludes the paper with a discussion of the main findings, limitations, and directions for future research.

\section{Related Work}\label{sec2}

At present, OM approaches can be broadly categorised into four main types: traditional knowledge-based systems, machine learning-based systems, pre-trained language model-based systems and more recently, large language models (LLMs)-based systems. \textbf{Traditional systems}, such as LogMap  and AML, rely primarily on lexical similarity, structural heuristics, and external resources like UMLS or WordNet \cite{anam2015review}. LogMap \cite{jimenez2011logmap} extracts class names and searches for matches via external lexicons while addressing logical inconsistencies by selecting alignments with higher confidence scores. AML \cite{faria2013agreementmakerlight} employs multiple matching strategies, including exact and character-based matchers, and has shown strong performance on medical datasets. However, traditional ontology matching systems largely rely on string-based similarity measures and curated external lexicons, which have been shown to limit their coverage when semantically equivalent concepts differ substantially at the lexical level~\cite{he2022bertmap}. 
As a result, such approaches may fail to identify valid correspondences that are not explicitly recorded in external dictionaries or lexicons.

With the rise of \textbf{machine learning}, several approaches have emerged that regard alignment as a classification or (embedding-based) similarity learning task \cite{doan2004ontology,kolyvakis-etal-2018-deepalignment,nkisi2018ontology,wang2018ontology,chen2021augmenting,bento2020ontology}. For example, DeepAlignment \cite{kolyvakis-etal-2018-deepalignment} vectorises class names and computes Euclidean distances to assess similarity, while the CNN-based system \cite{bento2020ontology} uses character-level embeddings and hierarchical context to train binary classifiers for equivalence detection. 

In parallel, \textbf{encoder-based pre-trained language models} such as BERT~\cite{devlin-etal-2019-bert} have been explored for ontology matching, leveraging contextualised representations learned from large-scale corpora. BioSTransformers \cite{menad2023biostransformers} adopt a Siamese architecture based on domain-specific BERT models to compute semantic similarity between biomedical concepts.  BERTMap \cite{he2022bertmap} fine-tunes a domain-specific BERT model on ontology alignment corpora mainly automatically extracted from synonyms, enabling it to capture subtle semantic differences even when lexical overlap is low. This approach has demonstrated strong results, particularly in biomedical applications.
BERTSubs \cite{DBLP:journals/www/ChenHGJDH23} takes a similar architecture as BERTMap but focuses on the subsumption relationship. 

More recently, the emergence of generative \textbf{LLMs}, including the GPT series, the Llama series, and the T5 series, has stimulated a new line of research in ontology alignment. 
These LLMs provide strong contextual reasoning and linguistic generalisation capabilities, making them attractive tools for capturing semantic relations that go beyond lexical similarity. 
A number of studies have explored prompting strategies and retrieval-augmented pipelines to exploit these capacities. 
For example, Norouz et~al.~\cite{Norouzi2023ConversationalOA} employed GPT-4o in a prompt-driven alignment workflow, reporting strong recall, while highlighting challenges in reliably distinguishing equivalence from hierarchical relations. 
Yuan et~al.~\cite{he2023exploring} evaluated both open- and closed-source LLMs on biomedical alignment tasks, incorporating structural cues to strengthen predictive reliability. 
Hybrid frameworks that combine embedding-based similarity retrieval with LLM verification have also been proposed~\cite{giglou2024llms4om,10.14778/3712221.3712222}, aiming to mitigate hallucination and improve alignment robustness. 
The Olala system~\cite{hertling2023olala} further integrates embedding-based filtering and post-processing modules with Llama-2, which is responsible for generating final alignment decisions. Hu et~al.~\cite{hu2025matching} have explored enriching ontological information by leveraging LLMs to expand or clarify incomplete concept metadata. 
Their approach employs LLaMA-3.3-70B-Instruct to, for instance, transform abbreviated forms into their likely full expressions and to generate explanatory descriptions for concepts with limited textual context. The latest version of LogMap-LLM~\cite{lushnei2025large} integrates GPT-4-mini and models from the Gemini family to assist in resolving uncertain and ambiguous candidate mappings. 
This hybrid integration enables the system to refine difficult alignment cases that conventional lexical- or structure-based heuristics struggle to disambiguate, resulting in modest improvements over the original LogMap pipeline.

Despite recent advances, existing LLM-based and LLM-augmented ontology matching approaches continue to face notable limitations. A central challenge arises from the fragmented and often sparse lexical information available within most ontologies~\cite{hertling2023olala, giglou2024llms4om, Norouzi2023ConversationalOA}. LLMs must rely heavily on labels, short synonyms, and minimally structural cues, which provide insufficient semantic context for distinguishing between closely related or hierarchically similar medical concepts. As a result, current LLM-based systems often struggle in fine-grained disambiguation scenarios~\cite{giglou2024llms4om}, where subtle terminological or relational differences must be accurately interpreted.

To address this limitation, our approach introduces an LLM-driven definition generation module that enriches each concept with a coherent, semantically informative description prior to alignment. These enriched definitions provide the LLM with a more holistic understanding of the underlying concepts, enabling more reliable and fine-grained mapping judgements compared with relying solely on the ontology’s native lexical fragments.

\section{Methodology}\label{sec3}
\subsection{Task Formulation}

OM is the process of discovering semantic correspondences between entities that belong to different ontologies. 
Formally, \rev{the input of the OM task consists of two ontologies,} the \textit{source ontology} $O_{\text{s}}$ and the \textit{target ontology} $O_{\text{t}}$. Let $C_{\text{s}}$ and $C_{\text{t}}$ denote the sets of named concepts contained in $O_{\text{s}}$ and $O_{\text{t}}$, respectively. 
The goal of OM is to identify a set of mappings that link semantically related concepts across the two ontologies. 
Each mapping consists of a concept pair $(c_{\text{s}}, c_{\text{t}})$, where $c_{\text{s}} \in C_{\text{s}}$ and $c_{\text{t}} \in C_{\text{t}}$. 
\rev{The output of the task is therefore a set of correspondences:}
\begin{equation}
M = \left\{ (c_{\text{s}}, c_{\text{t}}) \mid c_{\text{s}} \in C_{\text{s}},\, c_{\text{t}} \in C_{\text{t}}\right\}
\end{equation}

In practical settings, the OM task can encompass various types of semantic relations, such as \textit{equivalence}, \textit{subsumption}, and \textit{relatedness}, depending on the intended application or evaluation protocol. 
Accordingly, the OM task can be regarded as a process that bridges heterogeneous conceptual representations and enables interoperability of knowledge across ontologies.

In this study, the focus is restricted to the \textit{equivalence} problem, where the objective is to determine concept pairs that denote an identical meaning across the source and target ontologies.

\subsection{The Workflow of the GenOM Framework}

\begin{figure}
    \centering
    \includegraphics[width=5.1in]{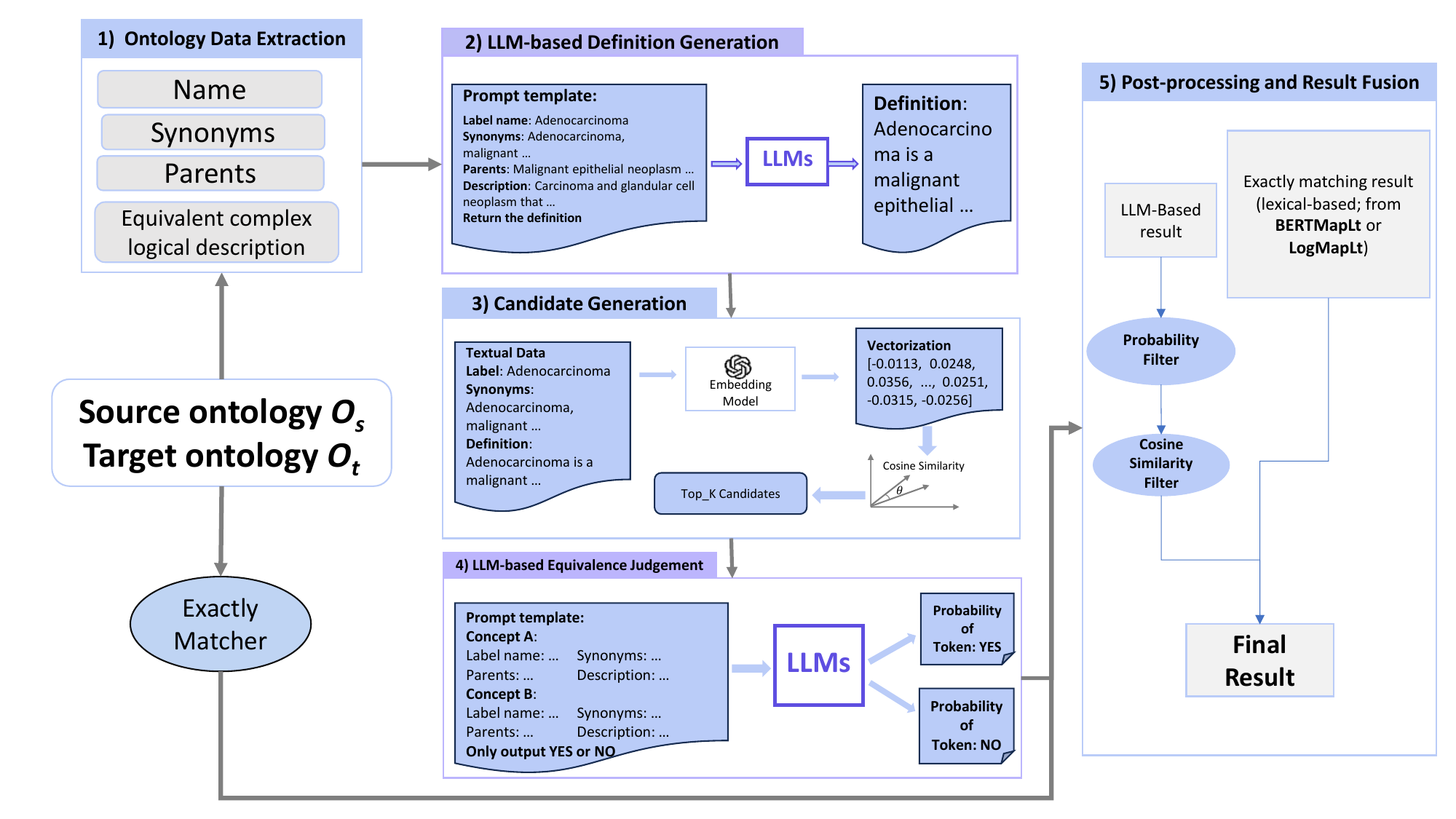}
    \caption{The Architecture of GenOM}
    \label{figure 1}
\end{figure}

The overall workflow of GenOM, illustrated in Figure~\ref{figure 1}, comprises five main components. 
1)~\textbf{Ontology data extraction} collects structural and lexical information for each named concept in the source and target ontologies, including labels, synonyms, parent concepts, and equivalence axioms. \rev{This information forms the foundational representation of each concept and serves as the input for the subsequent definition generation stage.}
2)~An \textbf{LLM-based definition generation} module produces natural language definitions based on paraphrased descriptions of concepts using the extracted information. This enriches their semantic representations—particularly for entities without explicit textual descriptions~\cite{10.1145/3746252.3761533,liu_are_2026}. \rev{The generated definitions provide richer contextual signals that are later used to compute semantic similarity between concepts.}
3)~A \textbf{candidate generation} stage employs an embedding-based retrieval process to compute pairwise cosine similarities between source and target concepts \rev{using their enriched textual representations. Based on these similarity scores, the system retrieves the top-$k$ candidate mappings for each source concept, which are then passed to the next stage for equivalence judgement.}
4)~An \textbf{LLM-based equivalence judgement} module evaluates each candidate pair, determining whether the two concepts convey equivalent meanings by leveraging their generated definitions and structural context. \rev{This stage performs semantic reasoning to verify the retrieved candidate mappings and produces equivalence predictions.}
5)~A \textbf{post-processing and result fusion} stage integrates the LLM prediction scores with cosine similarities and exact lexical matching results, retaining only high-confidence correspondences. \rev{This final step combines complementary signals from semantic reasoning, embedding similarity, and lexical matching to produce the final set of ontology mappings.}

Through the integration of definition generation, similarity retrieval, and semantic reasoning, GenOM combines the complementary strengths of embedding-based similarity, lexical precision, and LLM-based contextual understanding to achieve robust and scalable ontology matching. 


\paragraph{1)Ontology Data Extraction}
Both lexical and structural characteristics of each concept from the source and target ontologies are extracted and utilized to support the following alignment process. Specifically, the extracted information includes the concept's \textit{label} (defined by the annotation property \textit{rdfs:label}), a set of \textit{synonyms} (retrieved using the annotation properties listed in Table~\ref{tab:annotation-properties}), and its immediate \textit{parent concepts} as defined by the \textit{rdfs:subClassOf} relation. \rev{If certain types of information were unavailable for a concept, such as synonyms or parent concepts, the corresponding field was left empty.}

For concepts defined using \textit{EquivalentClass} axioms, the built-in \texttt{verbalisation} module in DeepOnto \cite{DBLP:journals/semweb/HeCDHAKS24} is used to convert logical expressions into natural language descriptions (Table~\ref{tab:dl_verbalization_vertical}). \rev{All extraction procedures were implemented using the DeepOnto framework.}

\begin{table}[h]
\centering
\begin{tabular}{ll}
\toprule
\textbf{Property} & \textbf{IRI} \\
\midrule
Label   & \texttt{http://www.w3.org/2000/01/rdf-schema\#label} \\
Synonym & \texttt{http://www.geneontology.org/formats/oboInOwl\#hasSynonym} \\
        & \texttt{http://www.geneontology.org/formats/oboInOwl\#hasExactSynonym} \\
        & \texttt{http://www.ebi.ac.uk/efo/alternative\_term} \\
        & \texttt{http://www.orpha.net/ORDO/Orphanet\_\#symbol} \\
        & \texttt{http://purl.org/sig/ont/fma/synonym} \\
        & \texttt{http://www.w3.org/2004/02/skos/core\#altLabel} \\
        & \texttt{http://www.w3.org/2004/02/skos/core\#prefLabel} \\
        & \texttt{http://ncicb.nci.nih.gov/xml/owl/EVS/Thesaurus.owl\#P108} \\
        & \texttt{http://ncicb.nci.nih.gov/xml/owl/EVS/Thesaurus.owl\#P90} \\
\bottomrule
\end{tabular}
\caption{Annotation property IRIs used for label and synonym extraction}
\label{tab:annotation-properties}
\end{table}

\begin{table}[htbp]
\centering

\begin{tabular}{|p{12.5cm}|}
\hline
\multicolumn{1}{|l|}{\textbf{\textit{EquivalentClass} Axiom in Description Logic}} \\
\hline

\textit{Product containing only betamethasone and calcipotriol (medicinal product)} $\equiv$ \\
\textit{MedicinalProduct} $\sqcap$ 
$\exists$ \textit{RoleGroup}.($\exists$ \textit{hasActiveIngredient}.\textit{Betamethasone}) $\sqcap$ \\
$\exists$ \textit{RoleGroup}.($\exists$ \textit{hasActiveIngredient}.\textit{Calcipotriol})\\
\hline
{\textbf{Verbalized Description by DeepOnto Toolkit}} \\
\hline
Medicinal product (product) that Role group (attribute) something that Has active ingredient (attribute) Betamethasone (substance) and something that Has active ingredient (attribute) Calcipotriol (substance). \\
\hline
\end{tabular}
\caption{Example of \textit{EquivalentClass} axiom, and the verbalized description of the logical expression}
\label{tab:dl_verbalization_vertical}
\end{table}

\paragraph{2)Definition Generation}

LLMs encode extensive general purpose and domain knowledge. They are leveraged and integrated with concept information explicitly represented in the ontology to enhance the semantic representation of the concept. In GenOM, definitions for medical concepts are generated by prompting an LLM with background information extracted from the ontology, including the concept's \textit{label}, \textit{synonyms}, \textit{parent concepts}, and, where applicable, the verbalised descriptions of logical expressions from \textit{EquivalentClass} axioms. 

Table~\ref{tab:prompt-templates} presents the prompt template employed for definition generation. In this template, the LLM is provided with both the available concept-specific information and the name of the source ontology. This additional context is intended to provide the LLM with high-level domain cues, which may facilitate the generation of more contextually appropriate definitions. The ontology name serves as supplementary guidance rather than an explicit source of supervision, and is particularly useful when concept-level annotations are sparse.

The amount and richness of information associated with a concept can vary considerably across ontologies. Some concepts are well-described, including multiple synonyms, hierarchical structure, and even formal axioms. In contrast, other concepts may be sparsely defined, often limited to a label with little or no supporting context. In such cases, the LLM’s internalised knowledge becomes essential in compensating for missing semantics.


\begin{table}[ht]
\centering
\begin{tabular}{|p{12.5cm}|}
\hline
\multicolumn{1}{|c|}{\textbf{Prompt Templates for Concept Definition Generation}} \\

\hline
\small
\textbf{Role: System} \\
You are generating a definition for a concept from the \texttt{\{$O_{\text{s}}$ name\}} ontology. The definition will be used to align it with candidate concepts in the \texttt{\{$O_{\text{t}}$ name\}} ontology.

You are a biomedical ontology expert. Your task is to generate a concise, alignment-friendly definition for a given biomedical concept. The definition should be semantically precise, distinguishable from related terms, and suitable for matching across ontologies.

Only return the definition. \\
\hline
\textbf{Role: User} \\
\hspace{1em} Concept: Product containing only betamethasone and calcipotriol (medicinal product)\\
\hspace{1em} Synonyms: Betamethasone and calcipotriol only product\\
\hspace{1em} Parents: Product containing betamethasone and calcipotriol (medicinal product)\\
\hspace{1em} Description: Medicinal product (product) that Role group (attribute) something that Has active ingredient (attribute) Betamethasone (substance) and something that Has active ingredient (attribute) Calcipotriol (substance)\\

\hline
\end{tabular}
\caption{The prompt template used for LLM-based definition generation.}
\label{tab:prompt-templates}
\end{table}

\paragraph{3)Candidate Mapping Generation}

To generate candidate concept pairs for alignment, GenOM adopts an embedding-based retrieval strategy. Concepts from both the source ontology $O_{\text{s}}$ and the target ontology $O_{\text{t}}$ are first encoded into fixed-size vector representations. Each concept is represented using a combination of its \textit{label}, \textit{synonyms}, and its enriched \textit{definition} produced in the previous step. Figure~\ref{fig:embedding_input} shows an example of the input text fed into the embedding model. Structural information such as hierarchical relations is deliberately excluded at this stage to reduce complexity in embedding. The framework adopts the \texttt{text-embedding-3-small}\footnote{\url{https://platform.openai.com/docs/models/text-embedding-3-small}} model provided by OpenAI for embedding generation. Compared to traditional encoder-based models such as Sentence-BERT \cite{reimers2019sentence}, this embedding model benefits from large-scale pre-training on diverse textual corpora and is better suited to capturing nuanced semantic differences expressed in longer, definition-enriched inputs, which is particularly important for distinguishing fine-grained biomedical concepts.

After generating embeddings for all target ontology concepts, we build a Hierarchical Navigable Small World (HNSW) index using the FAISS library \cite{johnson2019billion}. The index is constructed in the cosine similarity space by normalising all vectors to unit length, allowing FAISS to treat cosine similarity as equivalent to inner-product distance. This HNSW structure enables efficient approximate nearest-neighbour retrieval 
with sublinear query complexity, substantially reducing the computational cost compared with exhaustive pairwise similarity computation.

Once the index is constructed, each source concept embedding is queried against the FAISS HNSW index to retrieve its top-$k$ most similar target concepts under cosine similarity. This indexing-based retrieval mechanism significantly narrows the search space and provides a compact set of candidate mappings, which are subsequently passed to the LLM-based mapping judgement component.


\begin{figure}[ht]
\centering
\begin{tcolorbox}[title=Concept Representation Template, colback=gray!5, colframe=gray!80, width=0.9\textwidth]
\textbf{Label:} \{label\_string\} 
\textbf{Synonyms:} \{synonym\_1\}; \{synonym\_2\}; ... ; \{synonym\_n\} 
\textbf{Definition:} \{enriched\_definition\_generated\_by\_LLM\}
\end{tcolorbox}

\vspace{0.6em}

\begin{tcolorbox}[title=Embedding Input Example, colback=gray!5, colframe=gray!80, width=0.9\textwidth]
\textbf{Label:} Product containing only betamethasone and calcipotriol (medicinal product) 
\textbf{Synonyms:} Betamethasone and calcipotriol only product 
\textbf{Definition:} A medicinal product specifically formulated to contain solely 
betamethasone and calcipotriol as its active ... 
\end{tcolorbox}

\caption{Template and example of concept information used for embedding-based representation.}
\label{fig:embedding_input}
\end{figure}

\paragraph{4)LLM-based Equivalence Judgement}

In this stage, an LLM is employed to determine whether each candidate concept pair represents a semantic equivalence. Rather than prompting the model to generate full descriptive justifications, which would be time-consuming and potentially verbose, a lightweight classification strategy is adopted. Specifically, each concept pair is presented via a prompt designed to elicit a binary response — \texttt{YES} if the concepts are equivalent, and \texttt{NO} otherwise.
To support this, a prompt (as shown in Table \ref{tab:binary_prompt}) is constructed with strong instructional guidance, encouraging the model to respond using only a single classification token. 


The predicted equivalence score is computed using the probability assigned to the \texttt{YES} token, extracted directly from the model’s output logits. This follows the probabilistic interpretation of neural network outputs widely adopted in modern confidence estimation research, where token-level softmax probabilities are treated as calibrated or approximately calibrated confidence indicators \cite{guo2017calibration,kadavath2022language,si2022prompting}. 

Concretely, since the mapping judgement is explicitly formulated as a binary decision, we restrict the probability computation to the set of valid output labels $\mathcal{Y} = \{\texttt{YES}, \texttt{NO}\}$. Given the model’s output logits $z_{\texttt{YES}}$ and $z_{\texttt{NO}}$ at the final decoding position, the probability of semantic equivalence is computed as:

\begin{equation}
P(\texttt{YES}) = \text{softmax}(\mathbf{z})_{\texttt{YES}} = 
\frac{\exp(z_{\texttt{YES}})}{\exp(z_{\texttt{YES}}) + \exp(z_{\texttt{NO}})}
\end{equation}
Here, $z_{\texttt{YES}}$ denotes the logit corresponding to the token \texttt{YES}. The probability $P(\texttt{YES})$ then serves as the model’s confidence in semantic equivalence for the given concept pair. Concept pairs with $P(\texttt{YES})$ exceeding a predefined threshold are retained for alignment. 

Using token-level likelihoods as confidence estimates has been shown to correlate with model correctness in controlled-output settings, particularly when the answer space is restricted to a small set of discrete labels \cite{kadavath2022language,si2022prompting}. In our case, the mapping judgement is explicitly formulated as a binary decision (\texttt{YES}/\texttt{NO}), making token probability a natural and computationally efficient confidence proxy. This probability-based scoring approach substantially reduces inference time, eliminates the need for generating full-length textual responses, and provides a straightforward decision rule while maintaining high alignment precision.


\begin{table}[ht]
\centering
\begin{tabular}{|p{0.95\linewidth}|}
\hline
\textbf{Prompt Template for Equivalence Judgement} \\
\hline
\textbf{System Message:} \\
You are an expert in biomedical concept classification. You will be given two biomedical concepts. Based on the information provided, determine whether the two concepts refer to the same real-world entity (ontology matching). Only respond with \texttt{YES} or \texttt{NO}. \\
\hline
\textbf{User Message:} \\
\textbf{Concept A} \\
\hspace{1em} \textbf{Name:} \{\textit{lateral rectus nerve}\} \\
\hspace{1em} \textbf{Synonyms:} \{\textit{abducens nerve ...}\} \\
\hspace{1em} \textbf{Superclass:} \{\textit{peripheral nerve of head and neck (body structure) ...}\} \\
\hspace{1em} \textbf{Definition:} \{\textit{the lateral rectus nerve, also known as ...}\} \\[1ex]
\textbf{Concept B} \\
\hspace{1em} \textbf{Name:} \{\textit{abducent nerve [vi]}\} \\
\hspace{1em} \textbf{Synonyms:} \{\textit{nervus abducens ...}\} \\
\hspace{1em} \textbf{Superclass:} \{\textit{right posterior crico-arytenoid ligament ...}\} \\
\hspace{1em} \textbf{Definition:} \{\textit{the abducent nerve [vi] is a branch of the cranial nerve vi that innervates ...']}\} \\
\hline
\end{tabular}
\caption{The prompt template used for binary equivalence classification between ontology concepts. The fields in the brackets in the user message are populated with structured ontology information.}
\label{tab:binary_prompt}
\end{table}

\paragraph{5)Post-processing and Result Fusion}

To ensure the quality and reliability of the final alignment output, we apply a post-processing stage to filter and refine the candidate correspondences generated by the LLM. 
This stage relies on two thresholds, $\lambda_{\text{prob}}$ and $\lambda_{\text{cs}}$, which control confidence-based and similarity-based filtering, respectively.

Specifically, a threshold $\lambda_{\text{prob}}$ is imposed on the token-level probability associated with the \texttt{YES} response produced by the LLM. 
Candidate pairs whose confidence scores fall below this threshold are discarded. 
In parallel, cosine similarity scores obtained during candidate generation are considered, and pairs with embedding similarity lower than $\lambda_{\text{cs}}$ are removed to prevent semantically distant matches from being retained.

After this dual-filtering step, to further enhance precision, the remaining LLM outputs are merged with the results of two exact matching systems, \texttt{LogMapLt}\footnote{\url{https://github.com/ernestojimenezruiz/logmap-matcher}} and \texttt{BERTMapLt}\footnote{\url{https://github.com/KRR-Oxford/DeepOnto/tree/main/src/deeponto/align/bertmap}}. 
These systems are lightweight variants of their original models and are used here solely for string-based matching.

The detailed configuration of $\lambda_{\text{prob}}$ and $\lambda_{\text{cs}}$ are further analysed in Sections \ref{sec4} and \ref{sec6}.

\section{Overall System Performance Evaluation}\label{sec4}

\ignore{
This chapter is organised into six sections.

\begin{itemize}
    \item \textbf{Section~\ref{sec:dataset}} introduces the experimental datasets and evaluation metrics.
    \item \textbf{Section~\ref{sec:setup}} describes the models and algorithms, along with their hyperparameter configurations.
    \item \textbf{Section~\ref{sec:result}} reports the overall alignment results and benchmarks the proposed framework against state-of-the-art systems.
    
    \item \textbf{Section~\ref{sec:def_evl}} evaluates the definition generation component, detailing the assessment method and presenting definition-quality results obtained from various LLMs.

    \item \textbf{Section~\ref{sec:dif_llm}} analyses the impact of model scale by comparing LLMs of different parameter sizes.
    \item \textbf{Section~\ref{sec:ab_study}} presents ablation studies that assess the contribution of each component of the framework.
\end{itemize}
}
\ignore{
\subsection{Overall System Performance}
}
\subsection{Datasets and Evaluation Metrics}\label{sec:dataset}

The experiments were conducted on the \texttt{Bio-ML}\footnote{\url{https://krr-oxford.github.io/OAEI-Bio-ML/2024/index.html}} track \cite{he2022machine} of the Ontology Alignment Evaluation Initiative (OAEI) in 2024, which provides benchmarks specifically designed for biomedical ontology alignment, with a particular focus on machine learning-based systems. This track comprises five sub-tasks involving six widely used biomedical ontologies: \texttt{Systematized Nomenclature of Medicine Clinical Terms (SNOMED-CT)}, \texttt{National Cancer Institute Thesaurus (NCIT)}, \texttt{Foundational Model of Anatomy (FMA)}, \texttt{Human Disease Ontology (DOID)}, \texttt{Orphanet Rare Disease Ontology (ORDO)}, and \texttt{Online Mendelian Inheritance in Man (OMIM)}. \rev{These tasks cover multiple biomedical domains, including diseases, anatomical structures, pharmaceuticals, and neoplasms. As summarized in Table~\ref{tab:dataset_stats}, the five tasks correspond to different ontology pairs with varying semantic scopes and structural characteristics. All involved ontologies are large-scale biomedical ontologies containing thousands to tens of thousands of concepts, which makes the alignment problem particularly challenging. Evaluating ontology matching systems on such large and heterogeneous ontologies enables a more comprehensive assessment of the scalability and robustness of the proposed approach.}

\begin{table}[t]
\centering
\begin{tabular}{l l l r r r}
\hline
Source & Task & Category & \#SrcCls & \#TgtCls & \#Ref (equiv) \\
\hline
Mondo & OMIM--ORDO & Disease & 9,648 & 9,275 & 3,721 \\
Mondo & NCIT--DOID & Disease & 15,762 & 8,465 & 4,686 \\
UMLS & SNOMED--FMA & Body & 34,418 & 88,955 & 7,256 \\
UMLS & SNOMED--NCIT & Pharm & 29,500 & 22,136 & 5,803 \\
UMLS & SNOMED--NCIT & Neoplas & 22,971 & 20,247 & 3,804 \\
\hline
\end{tabular}
\caption{Statistics of the biomedical ontology matching tasks used in the experiments. \#SrcCls and \#TgtCls denote the number of source and target concepts, respectively, and \#Ref (equiv) indicates the number of reference equivalence mappings.}
\label{tab:dataset_stats}
\end{table}

The two official evaluation protocols of the Bio-ML track were adopted: \textit{global matching} which is to evaluate the system's ability to identify correct mappings among all the possible concept pairs across two ontologies, and \textit{local ranking} which focuses on ranking the correct target concept of a given source concept among a list of candidates. 

Precision (P), Recall (R) and F1-score are measured for global matching, and Mean Reciprocal Rank (MRR) and Hit@1 are calculated for local ranking. In the competition, these metrics provide a comprehensive assessment of the OM systems.

All evaluations strictly follow the official OAEI Bio-ML evaluation protocols and metric definitions, using the provided reference alignments, to ensure fair and comparable results across the systems.


\subsection{Experiment Setup}\label{sec:setup}

\rev{The implementation settings are organised according to the pipeline modules of GenOM, with each component corresponding to the stages described in Section~\ref{sec3}.}

\textbf{Definition Generation:} For semantic enrichment, \texttt{Qwen2.5-7B-Instruct}\footnote{\url{https://huggingface.co/Qwen/Qwen2.5-7B-Instruct}}, \texttt{Qwen2.5-14B-Instruct}\footnote{\url{https://huggingface.co/Qwen/Qwen2.5-14B-Instruct}}, \texttt{Qwen2.5-32B-Instruct}\footnote{\url{https://huggingface.co/Qwen/Qwen2.5-32B-Instruct}} and \texttt{Llama3.1-8B-Instruct}\footnote{\url{https://huggingface.co/meta-llama/Llama-3.1-8B-Instruct}} were utilised to generate concise definitions based on concept-level contextual information. For all definition generation experiments, we fixed the decoding hyperparameters to temperature = 0.7 and top\_p = 0.9 across all the LLMs. This setting corresponded to a moderately conservative sampling regime, which reduced excessive stochasticity in generation while still allowing limited variability beyond fully deterministic decoding. Such a configuration is suitable for definition generation, where outputs were expected to be stable and informative rather than highly diverse.

It is important to note that these hyperparameters were kept fixed throughout the experiments rather than tuned for individual models or tasks. This design choice allowed us to isolate the effect of the LLM itself on definition quality, avoided coupling the results to model-specific decoding behaviour, and ensured comparability across different models and experimental settings.


\textbf{Candidate Mapping Generation:} Cosine similarity computation and HNSW-based indexing were implemented via the \texttt{faiss} library to efficiently retrieve the top-$10$ most similar concepts for each source entity. 

\textbf{LLM-based Equivalence Judgement:} The  \texttt{Qwen2.5-32B-Instruct} LLM was applied to perform binary equivalence classification over the candidate pairs.  

\textbf{Post-processing and Result Fusion:} In the final stage, we applied two decision thresholds: the cosine similarity threshold ($\lambda_{\text{cs}}$) and the LLM-based probability threshold ($\lambda_{\text{prob}}$). To determine suitable values, we considered a predefined set of candidate thresholds $\{0.5, 0.6, 0.7, 0.8, 0.9\}$ with a step size of $0.1$. This range was chosen to focus on practically meaningful operating points and to exclude extreme high threshold values, which tend to produce overly conservative decisions and poor cross-task robustness in ontology matching. Threshold selection is performed on a single reference task, \textbf{SNOMED-NCIT (Neoplas)}. Based on its development results, a unified threshold value of $0.9$ was selected for both $\lambda_{\text{cs}}$ and $\lambda_{\text{prob}}$. These values were then fixed and applied unchanged to all remaining ontology matching tasks. 
Under such a setting, GenOM may not achieve the optimum performance on other tasks beyond SNOMED-NCIT (Neoplas), but we believe this setting is more consistent with real-world deployment and can reflect the framework's real performance especially on its robustness on different tasks.
In addition, results based on optimal thresholds obtained through task by task tuning are reported in Appendix~\ref{secA1}, for post-hoc analysis as a reference of GenOM's optimum performance and for comparison with other systems' results produced under this optimal setting.

With regard to the use of definition information, various LLMs were employed to generate textual definitions for medical concepts. However, in order to maintain experimental consistency, all components of the study that make use of definitions, except for the section specifically evaluating the definition quality produced by different LLMs, use a unified set of definitions generated by the Qwen2.5-7B-Instruct model.

We write \textbf{GenOM (Qwen32B)} for the GenOM framework using Qwen2.5-32B-Instruct as the LLM-based alignment model, i.e. the component responsible for predicting equivalence correspondences.

\subsection{Baseline Methods}
\rev{We compare GenOM with several representative ontology matching systems covering different methodological categories. These include traditional ontology matching systems, LogMap~\cite{jimenez2011logmap}, LogMapBio~\cite{jimenez2015logmap}, LogMapLt~\cite{jimenez2015logmap}, and Matcha; pre-trained language model (PLM)-based approaches including BERTMap~\cite{he2022bertmap}, BERTMapLt, and BioSTransMatch~\cite{menad2023biostransformers}; and recent LLM-based methods, Hu et al.~\cite{hu2025matching} (LLaMA-3.3-70B-Instruct), LLM4OM~\cite{giglou2024llms4om} (ChatGPT-3.5), and LogMap-LLM~\cite{lushnei2025large} (GPT-4o).}

\rev{Most baseline results are obtained from the official Bio-ML track website. In particular, LogMapLt and the BERTMap series were reimplemented within the track submission framework, while several other systems provided final alignment results without system reimplementation, likely using task-specific optimized configurations. The reported results for LLM4OM, LogMap-LLM, and Hu et al. were taken from their original publications. Among these systems, LLM4OM and Hu et al. use fixed thresholds across tasks, whereas LogMap-LLM relied on OpenAI and Gemini models without an explicit threshold parameter.}

\subsection{Overall Results}\label{sec:result}

\begin{table}[!ht]
\centering
\footnotesize
\renewcommand{\arraystretch}{0.7}
\begin{tabular}{llccccc}
\toprule
\textbf{Task} & \textbf{System} & \textbf{P} & \textbf{R} & \textbf{F1} & \textbf{MRR} & \textbf{H@1} \\

\midrule
\multirow{9}{*}{\textbf{SNOMED-NCIT (Neoplas)}} 
& LogMap        & 0.870 & 0.586 & 0.701 & NA    & NA    \\
& LogMapBio     & 0.784 & 0.795 & \underline{0.771} & NA    & NA    \\
& LogMapLt      & 0.951 & 0.517 & 0.670 & NA    & NA    \\
& Matcha        & 0.838 & 0.551 & 0.665 & 0.889 & \textbf{0.936} \\
& BERTMap       & 0.557 & 0.762 & 0.643 & 0.954 & 0.928 \\
& BERTMapLt     & 0.831 & 0.687 & 0.752 & 0.891 & 0.859 \\
& BioSTransMatch& 0.289 & 0.663 & 0.402 & 0.846 & 0.789 \\
& LLM4OM        & 0.470 & 0.530 & 0.495 & NA    & NA    \\
& LogMap-LLM      & 0.661 & 0.747 & 0.701 & NA    & NA    \\
& \textbf{GenOM(Qwen32B)} & 0.747 & 0.804 & \textbf{0.774} & \textbf{0.958} & \textbf{0.936} \\
\midrule
\midrule
\multirow{9}{*}{\textbf{SNOMED-NCIT (Pharm)}} 

& LogMap        & 0.966 & 0.607 & 0.746 & NA    & NA    \\
& LogMapBio     & 0.928 & 0.611 & 0.737 & NA    & NA    \\
& LogMapLt      & 0.996 & 0.599 & 0.748 & NA    & NA    \\
& Matcha        & 0.987 & 0.607 & \underline{0.752} & 0.936 & 0.921 \\
& BERTMap       & 0.971 & 0.585 & 0.730 & \textbf{0.969} & \textbf{0.951} \\
& BERTMapLt     & 0.981 & 0.574 & 0.724 & 0.849 & 0.773 \\
& BioSTransMatch& 0.584 & 0.844 & 0.690 & 0.943 & 0.918 \\
& LLM4OM        & 0.818 & 0.582 & 0.680 & NA    & NA    \\
& LogMap-LLM      & 0.855 & 0.621 & 0.719 & NA    & NA    \\
& \textbf{GenOM(Qwen32B)} & 0.957 & 0.622 & \textbf{0.754} & \underline{0.965} & \underline{0.943} \\

\midrule
\multirow{9}{*}{\textbf{SNOMED-FMA (Body)}} 
& LogMap        & 0.744 & 0.407 & 0.526 & NA    & NA    \\
& LogMapBio     & 0.827 & 0.577 & 0.680 & NA    & NA    \\
& LogMapLt      & 0.970 & 0.542 & 0.696 & NA    & NA    \\
& Matcha        & 0.887 & 0.502 & 0.641 & \textbf{0.950} & \textbf{0.935} \\
& BERTMap       & 0.979 & 0.662 & \textbf{0.790} & 0.944 & 0.920 \\
& BERTMapLt     & 0.979 & 0.655 & \underline{0.785} & 0.892 & 0.865 \\
& BioSTransMatch& 0.128 & 0.384 & 0.192 & 0.633 & 0.513 \\
& LLM4OM        & 0.211 & 0.326 & 0.256 & NA    & NA    \\
& LogMap-LLM      & 0.751 & 0.545 & 0.632 & NA    & NA    \\
& Hu et al. (2025)   & 0.231 & 0.279 & 0.253 & NA    & NA    \\
& \textbf{GenOM(Qwen32B)} & 0.902 & 0.674 & 0.771 & \underline{0.945} & \underline{0.926} \\

\midrule
\multirow{9}{*}{\textbf{OMIM-ORDO}} 
& LogMap        & 0.876 & 0.448 & 0.593 & NA    & NA    \\
& LogMapBio     & 0.866 & 0.609 & \underline{0.715} & NA    & NA    \\
& LogMapLt      & 0.940 & 0.252 & 0.397 & NA    & NA    \\
& Matcha        & 0.781 & 0.509 & 0.617 & 0.815 & 0.782 \\
& BERTMap       & 0.734 & 0.576 & 0.646 & \underline{0.880} & \underline{0.830} \\
& BERTMapLt     & 0.834 & 0.497 & 0.623 & 0.766 & 0.716 \\
& BioSTransMatch& 0.312 & 0.586 & 0.407 & 0.741 & 0.683 \\
& LLM4OM        & 0.718 & 0.580 & 0.641 & NA    & NA    \\
& LogMap-LLM      & 0.914 & 0.476 & 0.626 & NA    & NA    \\
& Hu et al. (2025)   & 0.725 & 0.540 & 0.619 & NA    & NA    \\
& \textbf{GenOM(Qwen32B)} & 0.834 & 0.643 & \textbf{0.726} & \textbf{0.930} & \textbf{0.904} \\

\midrule
\multirow{9}{*}{\textbf{NCIT-DOID}} 
& LogMap        & 0.934 & 0.668 & 0.779 & NA    & NA    \\
& LogMapBio     & 0.860 & 0.962 & \textbf{0.908} & NA    & NA    \\
& LogMapLt      & 0.983 & 0.575 & 0.725 & NA    & NA    \\
& Matcha        & 0.882 & 0.756 & 0.814 & 0.902 & 0.873 \\
& BERTMap       & 0.888 & 0.878 & 0.883 & \underline{0.959} & \underline{0.937} \\
& BERTMapLt     & 0.919 & 0.772 & 0.839 & 0.890 & 0.861 \\
& BioSTransMatch& 0.657 & 0.833 & 0.735 & 0.900 & 0.865 \\
& LLM4OM        & 0.862 & 0.801 & 0.830 & NA    & NA    \\
& LogMap-LLM      & 0.907 & 0.883 & 0.895 & NA    & NA    \\
& \textbf{GenOM(Qwen32B)} & 0.866 & 0.937 & \underline{0.901} & \textbf{0.967} & \textbf{0.950} \\
\bottomrule
\end{tabular}
\caption{Overall performance on the five OAEI 2024 Bio-ML tasks. NA indicates no results as the systems do not support the calculation of the metrics. Bold indicates the best performance, while underling indicates the second-best performance.}
\label{tab:bio_alignment_comparison}
\end{table}

Table~\ref{tab:bio_alignment_comparison} presented the performance of our method on the Bio-ML track in comparison with several state-of-the-art ontology alignment systems. Overall, GenOM (Qwen32B) exhibited competitive performance relative to other baseline systems on the Bio-ML track.

\begin{itemize}

\item On the \textbf{SNOMED–NCIT (Neoplas)} task, GenOM (Qwen32B) achieved the best results across all three evaluation metrics, namely F1-score, MRR, and H@1. In particular, its F1-score exceeded that of LogMapBio, a LogMap variant that explicitly integrated additional biomedical knowledge.

\item For the \textbf{SNOMED–NCIT (Pharm)} task, GenOM (Qwen32B) attained the highest F1-score among all evaluated systems. Although GenOM ranked second with respect to MRR and H@1, the absolute differences to the top-performing method, BERTMap, were small, amounting to 0.004 for MRR and 0.008 for H@1.

\item On the \textbf{SNOMED–FMA (Body)} task, GenOM (Qwen32B) ranked second in terms of both MRR and H@1, with marginal differences compared to the top-ranked system. While GenOM achieved the third-highest F1-score, the absolute gaps to the first- and second-ranked methods were limited. Specifically, BERTMap attained the highest F1-score of 0.790, whereas GenOM achieved 0.771. By contrast, the fourth-ranked system, LogMapLt, recorded an F1-score of 0.696. This corresponds to an absolute F1-score difference of 0.075 compared to LogMapLt, indicating that GenOM achieved performance comparable to the leading methods on this task.

\item On the \textbf{OMIM–ORDO} task, GenOM (Qwen32B) achieved the best performance across all three evaluation metrics. In particular, it outperformed the second-best system on ranking-based metrics, exceeding BERTMap by 0.05 in MRR and by 0.074 in H@1. This demonstrates a notable improvement in both metrics.

\item For the \textbf{NCIT–DOID} task, GenOM (Qwen32B) again achieved the highest scores on MRR and H@1. Although it ranked second in terms of F1-score, the absolute difference to the top-ranked system was only 0.007, further indicating that GenOM remained competitive on this task.

\end{itemize}

Compared with existing LLM-based approaches, GenOM (Qwen32B) achieved strong overall performance while relying on a model with a smaller parameter size. On the two Bio-ML tasks where Hu et al.~(2025)~\cite{hu2025matching} reported results (SNOMED--FMA and OMIM--ORDO), GenOM (Qwen32B) increased the F1-score by 0.518 and 0.107, respectively (averaging a 0.313 absolute gain across the two tasks). Across all five Bio-ML tasks, GenOM (Qwen32B) improved the mean F1-score over LLM4OM by 0.205 (35.3\%) and over LogMap-LLM by 0.071 (9.88\%).

In general, GenOM (Qwen32B) exhibited consistently strong and stable performance in different tasks and evaluation metrics. Even in tasks where it did not achieve the top rank, the performance gaps of the best-performing systems remained small, indicating that GenOM (Qwen32B) operated at a level comparable to state-of-the-art approaches. Furthermore, its strong performance on local ranking metrics demonstrated that GenOM (Qwen32B) was effective in distinguishing semantically similar biomedical concepts, highlighting its effectiveness in fine-grained semantic discrimination. \rev{Representative error cases of the LLM outputs are analysed in Appendix ~\ref{secC}.}

\section{Definition Generation Analysis}\label{sec5}

\begin{table}[t]
\centering
\small
\begin{tabular}{p{0.38\textwidth} p{0.52\textwidth}}
\toprule
\textbf{Concept} & \textbf{LLM-generated definition (label-like)} \\
\midrule
Middle cerebral artery infarction &
Infarction of the middle cerebral artery. \\[0.4em]
\midrule
Immune system organ benign neoplasm &
A benign neoplasm originating in an organ of the immune system. \\
\bottomrule
\end{tabular}
\caption{Examples of label-like definitions that largely restate the concept label without providing additional semantic information.}
\label{tab:label_like_examples}
\end{table}

In this section, we analyse the quality of the definitions generated by different LLMs during the \textbf{Definition Generation} stage. Several LLMs were used to produce definitions for the same set of biomedical ontology concepts, enabling a comparative study of their ability to capture domain semantics and support downstream ontology alignment.

Accurate and informative definitions play an essential role in ontology matching, particularly in the biomedical domain where many concepts differ only subtly~\cite{10.14778/3712221.3712222}.
The generated definitions in GenOM are expected to be factually accurate and medically informative, while being sufficiently \emph{discriminative} to differentiate the target concept from closely related alternatives.

In our experiments, we observed that definitions generated by different LLMs vary
substantially in quality, and that simple surface properties such as length or lexical
overlap provide limited insight into their quality. These observations motivate the need for an evaluation framework that can assess both the correctness of the generated content and its ability to discriminate between semantically similar concepts, while avoiding artificially high scores for trivial or label-like rewritings as illustrated in Table~\ref{tab:label_like_examples}.

\rev{For ease of reference, we use the following abbreviations: 
\textbf{Qwen7B} (Qwen2.5-7B-Instruct), 
\textbf{Qwen14B} (Qwen2.5-14B-Instruct), 
\textbf{Qwen32B} (Qwen2.5-32B-Instruct), 
and \textbf{Llama8B} (Llama3.1-8B-Instruct).}

\subsection{Hybrid Evaluation Method}

\ignore{
\begin{algorithm}
\caption{Hybrid Definition Quality Scoring Method}
\label{alg:def_scoring}
\begin{algorithmic}[1]

\Require Concept label $A$, generated definitions $Def = \{def_1, def_2, def_3, def_4\}$, 
nearest neighbouring concept $B$ (retrieved from the \emph{same} ontology), 
penalty coefficient $\lambda = 1$.

\Ensure Final adjusted scores $S = \{s_1', s_2', s_3', s_4'\}$.

\State \textbf{Single LLM Evaluation:} Query GPT-5-mini \emph{once}, providing
label $A$, nearest neighbour $B$, and the set of definitions $\{def_1,\dots,def_4\}$ 
in a single prompt. 
\Statex \hspace{0.8em}The LLM returns, for each $def_i$, a pair of scores:
\Statex \hspace{1.6em}discriminative score $d_i \in \{1,\dots,5\}$ (how well $def_i$ distinguishes $A$ from $B$)
\Statex \hspace{1.6em}correctness score $c_i \in \{1,\dots,5\}$ (medical / semantic correctness of $def_i$).

\For{$i = 1$ to $4$}
    \State \textbf{Normalised LLM Score:}
    \[
       s^{\text{norm}}_i = \frac{d_i + c_i}{10}
    \]

    \State \textbf{Jaccard Penalty:}
    \State Tokenise label $A$ into $t_A$, and definition $def_i$ into $t_i$
    \[
        J_i = \frac{|t_A \cap t_i|}{|t_A \cup t_i|}
    \]
    \If{$J_i > 0.5$}
        \State $P_i = \lambda \cdot (J_i - 0.5)$
    \Else
        \State $P_i = 0$
    \EndIf
    
    \State \textbf{Final Score:}
    \label{line:norm_score}
    \[
       s_i' = s^{\text{norm}}_i - P_i 
    \]
\EndFor

\State \Return $S = \{s_1', s_2', s_3', s_4'\}$

\end{algorithmic}
\end{algorithm}
}

\begin{algorithm}
\caption{Hybrid Definition Quality Scoring Method with Nearest-Neighbour Contrast}
\label{alg:def_scoring}
\begin{algorithmic}[1]

\Require Concept label $A$, generated definitions $Def = \{def_1, \dots, def_n\}$,
ontology $\mathcal{O}$ containing $A$,
SapBERT encoder $f(\cdot)$,
penalty coefficient $\lambda$ (set to $1$ in our experiments).

\Ensure Final adjusted scores $S = \{s_1',\dots, s_n'\}$.

\State \textbf{Nearest-Neighbour Selection (Contrastive Pair Construction):}
\Statex \hspace{0.8em}Encode the label of $A$ using SapBERT: $\mathbf{e}_A \leftarrow f(A)$.
\Statex \hspace{0.8em}For each candidate concept $x \in \mathcal{O}\setminus\{A\}$, compute $\mathbf{e}_x \leftarrow f(x)$ and 
similarity \Statex \hspace{0.8em}$\mathrm{sim}(\mathbf{e}_A,\mathbf{e}_x)$.
\Statex \hspace{0.8em}Select the nearest neighbour $B \leftarrow \arg\max_{x \in \mathcal{O}\setminus\{A\}} \mathrm{sim}(\mathbf{e}_A,\mathbf{e}_x)$.
\Statex \hspace{0.8em}\textit{Note:} $A$ and $B$ are retrieved from the same ontology $\mathcal{O}$, and the retrieval uses 
\Statex \hspace{0.8em}only lexical information (e.g., labels), not generated definitions.

\State \textbf{LLM-based Scoring:} Query GPT-5-mini \emph{once}, providing label $A$, nearest neighbour $B$, and the set of definitions $\{def_1,\dots,def_n\}$ in a single prompt (see Appendix~\ref{secB:prompt_template}).
\Statex \hspace{0.8em}The LLM returns, for each $def_i$, a pair of scores:
\Statex \hspace{1.6em}Discriminative score $d_i \in \{1,\dots,5\}$ (how well $def_i$ distinguishes $A$ from $B$).
\Statex \hspace{1.6em}Correctness score $c_i \in \{1,\dots,5\}$ (semantic correctness of $def_i$).

\For{$i = 1$ to $n$}
    \State \textbf{Normalised LLM Score:} $\displaystyle s^{\text{norm}}_i \leftarrow \frac{d_i + c_i}{10}$.

    \State \textbf{Jaccard-based Penalty:}
    \State Tokenise label $A$ into $T_A$, definition $def_i$ into $T_i$. Compute   $\displaystyle J_i \leftarrow \frac{|T_A \cap T_i|}{|T_A \cup T_i|}$.
    \If{$J_i > 0.5$}
        \State $P_i \leftarrow \lambda \cdot (J_i - 0.5)$
    \Else
        \State $P_i \leftarrow 0$
    \EndIf

    \State \textbf{Final Score:} $\displaystyle s_i' \leftarrow s^{\text{norm}}_i - P_i$. \label{line:norm_score}
\EndFor

\State \Return $S = \{s_1', \dots, s_n'\}$

\end{algorithmic}
\end{algorithm}

We employ \texttt{GPT-5-mini}\footnote{\url{https://platform.openai.com/docs/models/gpt-5-mini}}, a proprietary LLM with a larger parameter scale than the open-source LLMs used in GenOM, as an external evaluator for assessing the quality of generated definitions. The evaluator was not used in any stage of definition generation or ontology matching, thereby serving solely as an independent assessment component.

The proposed evaluation framework consists of three components: 
(i) Nearest-Neighbour Selection for constructing contrastive concept pairs, 
(ii) LLM-based Scoring of \textbf{correctness} and \textbf{discriminativeness}, and 
(iii) a Jaccard-based Penalty to reduce redundancy with the concept label. 
The complete procedure is summarised in Algorithm~\ref{alg:def_scoring}.

Concretely, for each concept $A$ and its set of generated definitions $\{def_1, \dots, def_n\}$, 
we first identify a lexically nearest neighbour $B$ from the same ontology using SapBERT~\cite{liu2021self} embeddings. 
This neighbour serves as a contrastive reference for evaluating the discriminative quality of definitions. 
The retrieval relies solely on lexical information (e.g., concept labels), and the nearest neighbour is selected as the top-1 most similar concept within the same ontology. 
Importantly, generated definitions are not used during this retrieval step, ensuring that the selection of $B$ is independent of generated definitions.

Next, the concept label $A$, its nearest neighbour $B$, and all candidate definitions are provided to \texttt{GPT-5-mini} in a single prompt (see Appendix~\ref{secB:prompt_template}). 
For each definition $def_i$, the LLM assigns two scores on a discrete scale from 1 to 5: 
a \emph{correctness} score $c_i$, reflecting semantic and biomedical validity, 
and a \emph{discriminativeness} score $d_i$, measuring how effectively $def_i$ distinguishes $A$ from $B$. 
These scores are combined into a normalised LLM score, as defined in Algorithm~\ref{alg:def_scoring}, yielding values in the range $[0,1]$ as a relative quality score.

To penalise definitions that closely resemble the concept label, 
we compute a Jaccard similarity $J_i$ between the tokenised label $A$ and the definition $def_i$. 
Both the label and the definition are tokenised at the word level, and common English stopwords (e.g., \emph{a}, \emph{an}, \emph{the}, \emph{of}, \emph{in}) are removed prior to computing the intersection and union. 
This design ensures that the redundancy penalty is driven by semantically meaningful lexical overlap rather than superficial matches on high-frequency function words. 
When $J_i > 0.5$, a penalty $P_i = \lambda (J_i - 0.5)$ is applied (with $\lambda = 1$ in our experiments); otherwise, the penalty is set to zero. The threshold of 0.5 was chosen as a conservative heuristic to penalise substantial lexical reuse, while allowing partial overlap that is often unavoidable in concise definitions. 
The final adjusted score is computed as $s_i' = s_i^{\text{norm}} - P_i$, where higher values indicate higher-quality definitions.

Due to the computational cost of \texttt{GPT-5-mini}, we randomly sampled 100 concepts from each ontology for evaluation. 
For each concept, the number of generated definitions was fixed to four in our experiments, corresponding to definitions produced by different LLMs. 
The resulting score $S$ provides a composite measure that integrates semantic correctness, discriminative capability, and redundancy penalisation.

Beyond the automated evaluation method described above, we also conducted a small-scale manual inspection as a qualitative sanity check to provide additional context for the proposed evaluation method. 
Details of this analysis and its limitations are reported in Appendix~\ref{secB:manual_check}.

\subsection{Results and Analysis for the Definition Evaluation Task}

\begin{table*}[t]
\footnotesize
\renewcommand{\arraystretch}{0.75}
\caption{LLM definition--generation performance across ontology subsets.
Each subset contained 100 randomly sampled concepts. For each concept, the
four LLMs independently generated a definition, and raw/final scores were
averaged over the 100 samples. \textit{Note.} Raw scores were obtained from the LLM-based assessment, and final scores incorporated the Jaccard-based redundancy penalty. Both scores were averaged over the 100 concepts per (task, ontology, LLM) combination.}
\label{tab:llm-all}
\centering
\begin{tabular*}{\textwidth}{@{\extracolsep{\fill}}lllcc}
\toprule
Task & Ontology & LLM & Avg.\ raw score & Avg.\ final score \\
     &          &     & (0--1)          & (0--1)            \\
\midrule
\multirow{8}{*}{Pharm}
  & \multirow{4}{*}{SNOMED} 
    & Qwen7B   & 0.768 & 0.752 \\
  &       & Llama8B   & 0.744 & 0.721 \\
  &       & Qwen14B  & \textbf{0.810} & \textbf{0.797} \\
  &       & Qwen32B  & 0.801 & 0.794 \\
\cmidrule(lr){2-5}
  & \multirow{4}{*}{NCIT} 
    & Qwen7B   & 0.713 & 0.713 \\
  &       & Llama8B   & 0.663 & 0.654 \\
  &       & Qwen14B  & \textbf{0.765} & \textbf{0.764} \\
  &       & Qwen32B  & 0.746 & 0.745 \\
\midrule
\multirow{8}{*}{Neoplas}
  & \multirow{4}{*}{SNOMED} 
    & Qwen7B   & 0.776 & 0.775 \\
  &       & Llama8B   & 0.745 & 0.707 \\
  &       & Qwen14B  & 0.806 & 0.805 \\
  &       & Qwen32B  & \textbf{0.813} & \textbf{0.809} \\
\cmidrule(lr){2-5}
  & \multirow{4}{*}{NCIT} 
    & Qwen7B   & 0.725 & 0.725 \\
  &       & Llama8B   & 0.685 & 0.675 \\
  &       & Qwen14B  & \textbf{0.785} & \textbf{0.784} \\
  &       & Qwen32B  & 0.753 & 0.753 \\
\midrule
\multirow{8}{*}{Body}
  & \multirow{4}{*}{SNOMED} 
    & Qwen7B   & 0.695 & 0.695 \\
  &       & Llama8B   & 0.652 & 0.609 \\
  &       & Qwen14B  & 0.676 & 0.674 \\
  &       & Qwen32B  & \textbf{0.712} & \textbf{0.708} \\
\cmidrule(lr){2-5}
  & \multirow{4}{*}{FMA} 
    & Qwen7B   & 0.601 & 0.601 \\
  &       & Llama8B   & 0.641 & 0.591 \\
  &       & Qwen14B  & 0.724 & 0.695 \\
  &       & Qwen32B  & \textbf{0.740} & \textbf{0.715} \\
\midrule
\multirow{8}{*}{Disease}
  & \multirow{4}{*}{NCIT}
    & Qwen7B   & \textbf{0.781} & \textbf{0.781} \\
  &       & Llama8B   & 0.678 & 0.658 \\
  &       & Qwen14B  & 0.752 & 0.750 \\
  &       & Qwen32B  & 0.762 & 0.761 \\
\cmidrule(lr){2-5}
  & \multirow{4}{*}{DOID}
    & Qwen7B   & 0.754 & 0.754 \\
  &       & Llama8B   & 0.634 & 0.626 \\
  &       & Qwen14B  & 0.755 & 0.750 \\
  &       & Qwen32B  & \textbf{0.805} & \textbf{0.804} \\

\midrule
\multirow{8}{*}{Disease}
  & \multirow{4}{*}{OMIM}
    & Qwen7B   & 0.743 & 0.742 \\
  &       & Llama8B   & 0.668 & 0.659 \\
  &       & Qwen14B  & 0.740 & 0.735 \\
  &       & Qwen32B  & \textbf{0.749} & \textbf{0.746} \\
\cmidrule(lr){2-5}
  & \multirow{4}{*}{ORDO}
    & Qwen7B   & 0.802 & 0.802 \\
  &       & Llama8B   & 0.725 & 0.721 \\
  &       & Qwen14B  & \textbf{0.806} & \textbf{0.806} \\
  &       & Qwen32B  & 0.804 & 0.803 \\
\bottomrule
\end{tabular*}

\end{table*}

Table~\ref{tab:llm-all} reported the average raw (LLM-based) and final (penalised) definition scores across ontology subsets, where the final score incorporated the Jaccard-based redundancy penalty. Three observations emerged.

First, the Qwen family consistently achieved higher final scores than Llama8B across the evaluated subsets. 
When averaging over all (task, ontology) subsets, Qwen32B and Qwen14B obtained higher mean final scores (0.764 and 0.756, respectively) than Qwen7B (0.734), while Llama8B yielded a substantially lower mean final score (0.662). 
Moreover, Qwen32B exhibited the strongest overall performance, achieving the best final score in five out of the ten subsets, with Qwen14B ranking first in four subsets. 
The only subset where a smaller model led was Disease--NCIT, where Qwen7B attained the highest final score (0.781).

Second, model scaling improved definition quality in general, but the gains were not strictly monotonic across subsets. 
Although Qwen32B achieved the strongest overall performance under the proposed evaluation, Qwen14B outperformed it on several individual subsets, including Pharm--SNOMED (0.797 vs.\ 0.794), Pharm--NCIT (0.764 vs.\ 0.745), Disease--ORDO (0.806 vs.\ 0.803), and Neoplas--NCIT (0.784 vs.\ 0.753). 
This showed that increasing model size did not consistently result in higher definition scores for every ontology subset.

Third, the redundancy penalty affected models to markedly different extents. 
Across subsets, Llama8B showed a substantially larger average drop from raw to final score (mean $\Delta \approx 0.021$) compared with Qwen models (Qwen14B: $\Delta \approx 0.006$; Qwen32B: $\Delta \approx 0.005$; Qwen7B: $\Delta \approx 0.002$). 
The largest drops for Llama8B were observed on the Body subsets and several SNOMED subsets (e.g., Neoplas--SNOMED: 0.745$\rightarrow$0.707; Body--SNOMED: 0.652$\rightarrow$0.609; Body--FMA: 0.641$\rightarrow$0.591), indicating that its outputs were affected more strongly by the Jaccard-based redundancy penalty under the same evaluation protocol.

Overall, these results suggested that larger Qwen models tended to produce higher-quality definitions under the proposed hybrid evaluation, with Qwen32B providing the most stable performance across subsets. 
At the same time, the performance gap between Qwen14B and Qwen32B remains small across most subsets, despite the general advantage of larger models. 
This pattern indicates that increasing model scale yields diminishing returns in definition quality, and suggest that model size, while influential, is not the sole factor determining final performance under this evaluation. 
The substantial difference between Llama8B and Qwen7B at a comparable parameter scale is consistent with this observation.

\section{Impact of different LLMs}\label{sec6}

\begin{figure}
    \centering
    \includegraphics[width=5in]{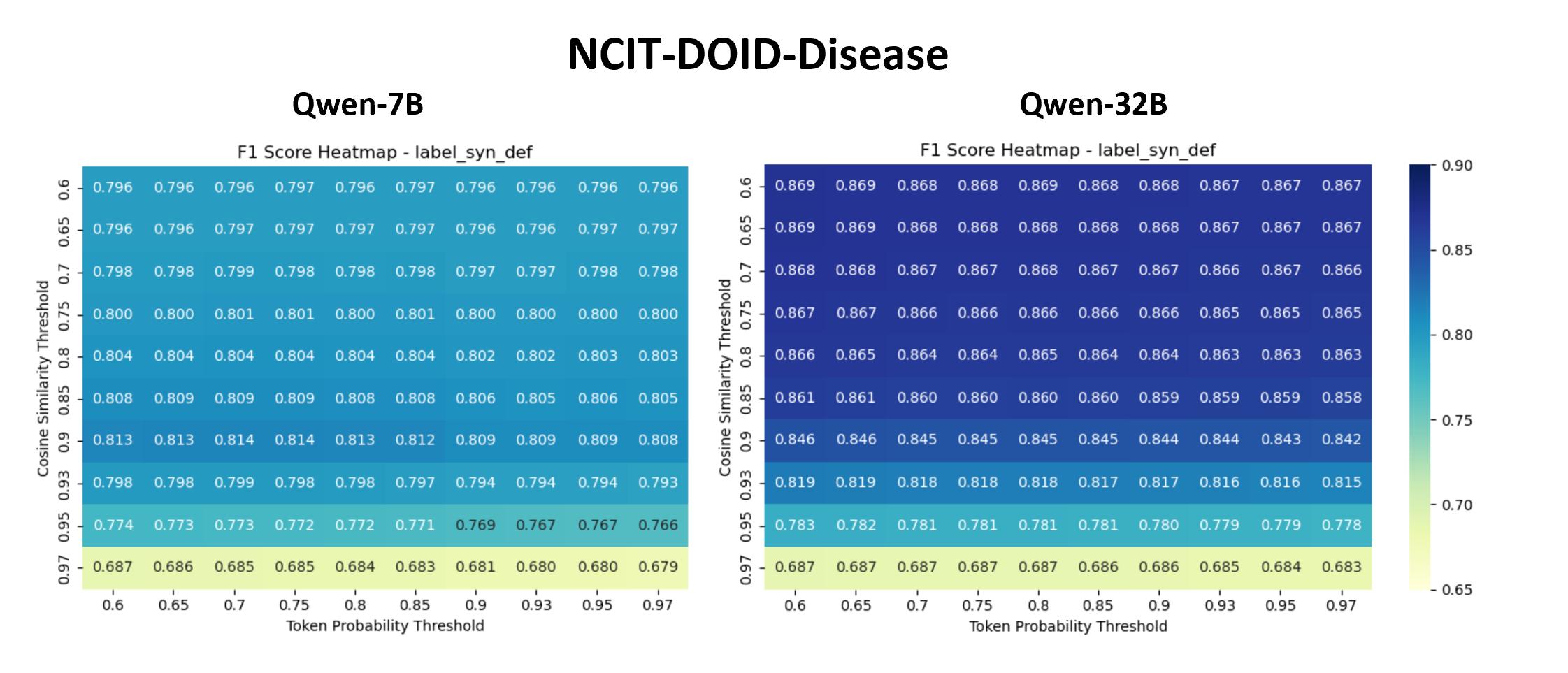}
    \includegraphics[width=5in]{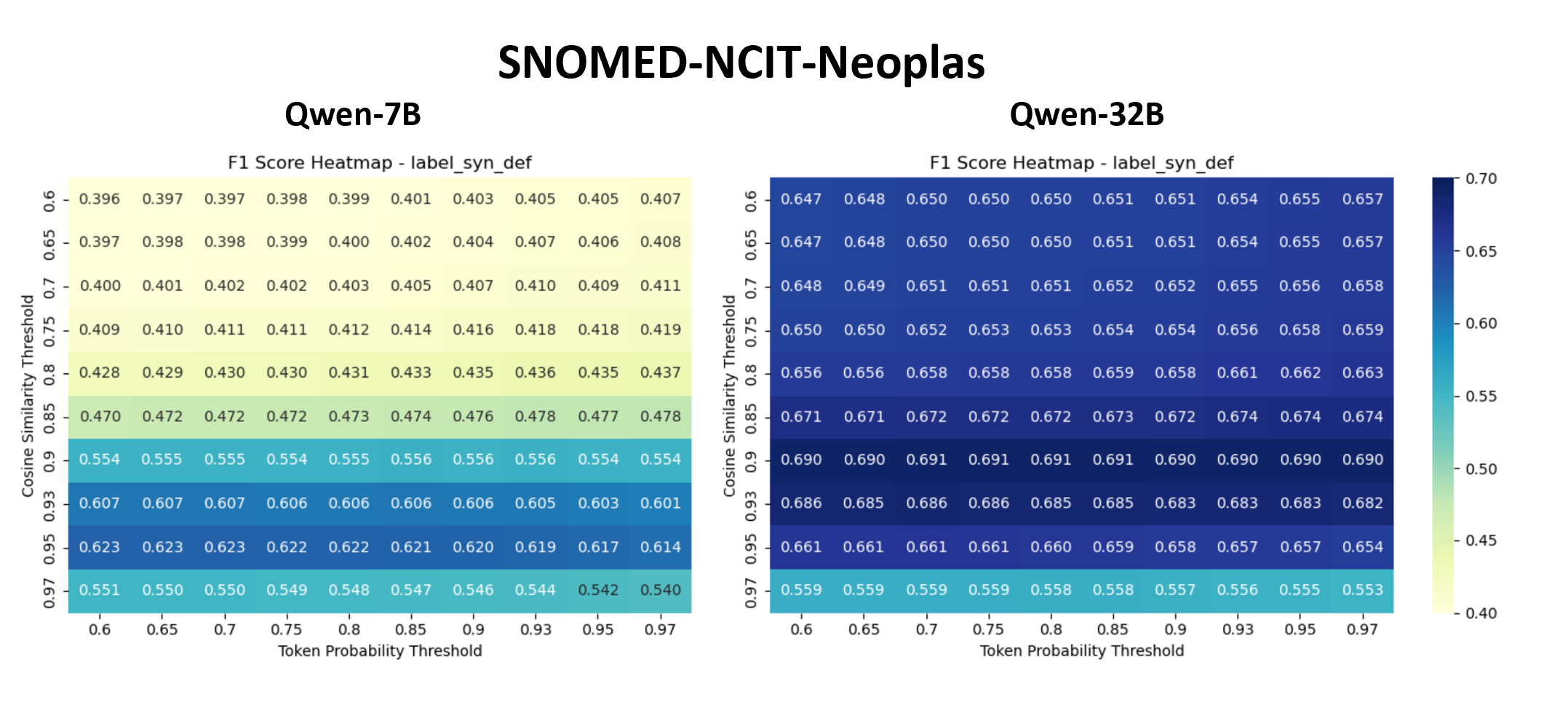}
    \caption{F1-Scores of GenOM under the LLM-only setting when the thresholds of cosine similarity and token probability are set to different values. Results of two matching tasks using Qwen7B and Qwen32B are reported.}
    \label{impact_llms}
\end{figure}

Figure~\ref{impact_llms} presents F1-score heatmaps for two representative tasks, NCIT--DOID (Disease) and SNOMED--NCIT (Neoplas), evaluated across a grid of the token-probability threshold and the cosine-similarity threshold for both Qwen7B and Qwen32B.  These results were based on the LLM-only setting, where the final alignment decisions were produced solely by the LLM without incorporating any exact-matching modules. Each candidate pair was evaluated using the labels, synonyms and generated definitions of the concepts, allowing us to directly compare the performance of different LLMs.

Across both tasks, Qwen32B consistently achieved higher F1 scores than Qwen7B, which is consistent with its larger model capacity. More importantly, the two models differ in their sensitivity to the cosine-similarity threshold.

For the NCIT--DOID task, both models showed very limited variation in F1 scores as the threshold changed across the examined range. This suggested that the alignment signals---derived from labels, synonyms, and generated definitions---were sufficiently strong and consistent that the equivalence-judgement stage of the LLM was largely insensitive to threshold selection. The underlying ontologies in this pair shared relatively homogeneous, clinically oriented concept representations, which made the equivalence classification problem comparatively straightforward even for the smaller model.

In contrast, the SNOMED-NCIT (Neoplas) task displayed substantial variability, particularly for Qwen7B.  

This phenomenon was not confined to the SNOMED–NCIT (Neoplas) task; it also appeared consistently in other tasks involving SNOMED. When SNOMED participated in the alignment pair, the smaller model became highly dependent on strict cosine-similarity thresholds: relaxing the threshold resulted in a rapid increase in false-positive equivalence predictions, leading to sharply reduced F1 scores. This instability can be attributed to the difficulty of performing cross-ontology equivalence judgement when one side (SNOMED) encoded highly granular and structurally complex concepts.

SNOMED terms frequently contained fine-grained morphologic, anatomical or 
structural information, e.g.~\emph{Adenolymphoma (morphologic abnormality)}, \emph{Dermatofibroma (disorder)}, \emph{Abducens nerve structure (body structure)}, whereas the corresponding concepts in NCIT were often expressed using more concise, high-level clinical terminology, e.g.~\emph{Recurrent Respiratory Papillomatosis}. Qwen7B struggled to reliably determine whether such asymmetric concept pairs should be treated as equivalent or merely related, producing many borderline cases that were filtered out only when strict cosine-similarity thresholds were applied.

By contrast, Qwen32B demonstrated far greater robustness across threshold settings. The larger model can better interpret SNOMED’s layered semantic structure and more reliably distinguish true equivalence from looser semantic relatedness, resulting in smoother heatmaps and reduced threshold sensitivity. These findings highlighted the intrinsic difficulty of aligning fine-grained, richly modelled ontologies such as SNOMED, and underscored the role of model capacity in achieving stable equivalence judgements under this setting.

\section{Ablation Study}\label{sec7}

\subsection{Impact of Definition Enrichment on Local Ranking and Candidate Retrieval}

We investigated the impact of definition-based semantic enrichment on two stages of the pipeline: LLM-based equivalence judgement and embedding-based candidate generation.

For LLM-based equivalence judgement, we compared a label-only setting with a label-plus-definition setting. The local ranking result was shown in Figure~\ref{fig:definition-impact-local-ranking}. In particular, both MRR and Hit@1 showed clear improvements in all tasks except for \textit{SNOMED-FMA (body)}, where performance remains comparable. \rev{Among these, the improvement is most pronounced in the \textit{OMIM–ORDO} task, where the performance gains reach 0.18 and 0.235 in terms of MRR and Hit@1, respectively.}

For the embedding-based candidate generation stage, we additionally reported Hit@5 and Hit@10, as these metrics are essential for determining the appropriate top-$k$ value—that is, how many candidate concepts should be passed to the LLM for equivalence judgement. As shown in Figure \ref{fig:embedding-definition-impact}, incorporating definition information led to improvements across all three metrics (Hit@1, Hit@5, Hit@10) in all tasks, except for a slight drop in Hit@10 on the \textit{SNOMED-NCIT (pharm)} task. \rev{It is also noteworthy that the inclusion of definitions leads to the most pronounced improvement in Hit@1, whereas the effect becomes considerably attenuated when the evaluation scope is extended to Hit@10, suggesting that definitions primarily enhance fine-grained semantic discrimination rather than broad candidate retrieval.} 

\rev{Overall, these results indicate that incorporating concept definitions improves performance at both stages of the pipeline. In the LLM-based equivalence judgement stage, the additional definition information helps the model better understand the semantics of concepts, which leads to more reliable alignment decisions. In the embedding-based candidate generation stage, definitions provide richer textual information that improves the quality of the retrieved candidates. Together, these findings show that concept definitions support both semantic comparison in the LLM stage and candidate retrieval in the embedding stage. More details can be found in the Appendix~\ref{secA2}}

\ignore{
\begin{table}[ht]
\centering
\footnotesize
\begin{tabular}{|l|c|c|}
\hline
\textbf{Task} & \textbf{MRR} & \textbf{Hit@1}  \\
\hline
OMIM\_ORDO (with) & 0.903  & 0.860 \\
OMIM\_ORDO (without) & 0.723 & 0.625 \\
NCIT\_DOID (with) & 0.949  & 0.924 \\
NCIT\_DOID (without) & 0.919  & 0.882 \\
SNOMED\_NCIT\_pharm (with) & 0.918  & 0.869 \\
SNOMED\_NCIT\_pharm (without) & 0.806  & 0.739 \\
SNOMED\_NCIT\_neoplas (with) & 0.928  & 0.888 \\
SNOMED\_NCIT\_neoplas (without) & 0.891 & 0.841 \\
SNOMED\_FMA\_body (with) & 0.871  & 0.819 \\
SNOMED\_FMA\_body (without) & 0.866  & 0.819 \\
\hline
\end{tabular}
\caption{Impact of concept definitions on LLM-based Local Ranking (Qwen32B)}
\label{tab:localranking-definition-impact}
\end{table}
}

\begin{figure}[ht]
\centering

\begin{subfigure}[t]{0.48\textwidth}
\centering
\begin{tikzpicture}
\begin{axis}[
    ybar,
    bar width=8pt,
    width=\textwidth,
    height=6cm,
    ymin=0.5, ymax=1.0,
    ylabel={MRR},
    symbolic x coords={
        OMIM--ORDO,
        NCIT--DOID,
        SNOMED--NCIT (pharm),
        SNOMED--NCIT (neoplas),
        SNOMED--FMA (body)
    },
    xtick=data,
    x tick label style={rotate=30, anchor=east},
    legend style={at={(0.5,1.05)},anchor=south,legend columns=2},
    legend image code/.code={
        \draw[#1, draw=none] (0cm,-0.1cm) rectangle (0.4cm,0.2cm);
    }
]
\addplot coordinates {
    (OMIM--ORDO, 0.903)
    (NCIT--DOID, 0.949)
    (SNOMED--NCIT (pharm), 0.918)
    (SNOMED--NCIT (neoplas), 0.928)
    (SNOMED--FMA (body), 0.871)
};
\addplot coordinates {
    (OMIM--ORDO, 0.723)
    (NCIT--DOID, 0.919)
    (SNOMED--NCIT (pharm), 0.806)
    (SNOMED--NCIT (neoplas), 0.891)
    (SNOMED--FMA (body), 0.866)
};
\legend{With definitions,Without definitions}
\end{axis}
\end{tikzpicture}
\caption{MRR}
\end{subfigure}
\hfill
\begin{subfigure}[t]{0.48\textwidth}
\centering
\begin{tikzpicture}
\begin{axis}[
    ybar,
    bar width=8pt,
    width=\textwidth,
    height=6cm,
    ymin=0.5, ymax=1.0,
    ylabel={Hit@1},
    symbolic x coords={
        OMIM--ORDO,
        NCIT--DOID,
        SNOMED--NCIT (pharm),
        SNOMED--NCIT (neoplas),
        SNOMED--FMA (body)
    },
    xtick=data,
    x tick label style={rotate=30, anchor=east},
]
\addplot coordinates {
    (OMIM--ORDO, 0.860)
    (NCIT--DOID, 0.924)
    (SNOMED--NCIT (pharm), 0.869)
    (SNOMED--NCIT (neoplas), 0.888)
    (SNOMED--FMA (body), 0.819)
};
\addplot coordinates {
    (OMIM--ORDO, 0.625)
    (NCIT--DOID, 0.882)
    (SNOMED--NCIT (pharm), 0.739)
    (SNOMED--NCIT (neoplas), 0.841)
    (SNOMED--FMA (body), 0.819)
};
\end{axis}
\end{tikzpicture}
\caption{Hit@1}
\end{subfigure}

\caption{Impact of concept definitions on LLM-based local ranking (Qwen32B).}
\label{fig:definition-impact-local-ranking}
\end{figure}

\begin{figure}[ht]
\centering

\begin{subfigure}[t]{0.45\textwidth}
\centering
\begin{tikzpicture}
\begin{axis}[
    ybar,
    bar width=6pt,
    width=\textwidth,
    height=4.5cm,
    ymin=0.6, ymax=1.0,
    ylabel={Hit@1},
    symbolic x coords={
        OMIM--ORDO,
        NCIT--DOID,
        SNOMED--NCIT (pharm),
        SNOMED--NCIT (neoplas),
        SNOMED--FMA (body)
    },
    xtick=data,
    x tick label style={rotate=35, anchor=east},
    legend style={at={(0.5,1.12)},anchor=south,legend columns=2},
    legend image code/.code={
        \draw[#1, draw=none] (0cm,-0.1cm) rectangle (0.4cm,0.2cm);
    },
]
\addplot coordinates {
    (OMIM--ORDO, 0.737)
    (NCIT--DOID, 0.886)
    (SNOMED--NCIT (pharm), 0.747)
    (SNOMED--NCIT (neoplas), 0.722)
    (SNOMED--FMA (body), 0.695)
};
\addplot coordinates {
    (OMIM--ORDO, 0.724)
    (NCIT--DOID, 0.786)
    (SNOMED--NCIT (pharm), 0.712)
    (SNOMED--NCIT (neoplas), 0.718)
    (SNOMED--FMA (body), 0.642)
};
\legend{With definitions,Without definitions}
\end{axis}
\end{tikzpicture}
\caption{Hit@1}
\end{subfigure}
\hfill
\begin{subfigure}[t]{0.45\textwidth}
\centering
\begin{tikzpicture}
\begin{axis}[
    ybar,
    bar width=6pt,
    width=\textwidth,
    height=4.5cm,
    ymin=0.6, ymax=1.0,
    ylabel={Hit@5},
    symbolic x coords={
        OMIM--ORDO,
        NCIT--DOID,
        SNOMED--NCIT (pharm),
        SNOMED--NCIT (neoplas),
        SNOMED--FMA (body)
    },
    xtick=data,
    x tick label style={rotate=35, anchor=east},
]
\addplot coordinates {
    (OMIM--ORDO, 0.879)
    (NCIT--DOID, 0.974)
    (SNOMED--NCIT (pharm), 0.942)
    (SNOMED--NCIT (neoplas), 0.896)
    (SNOMED--FMA (body), 0.912)
};
\addplot coordinates {
    (OMIM--ORDO, 0.875)
    (NCIT--DOID, 0.940)
    (SNOMED--NCIT (pharm), 0.911)
    (SNOMED--NCIT (neoplas), 0.896)
    (SNOMED--FMA (body), 0.883)
};
\end{axis}
\end{tikzpicture}
\caption{Hit@5}
\end{subfigure}

\vspace{0.8em}

\begin{subfigure}[t]{0.45\textwidth}
\centering
\begin{tikzpicture}
\begin{axis}[
    ybar,
    bar width=6pt,
    width=\textwidth,
    height=4.5cm,
    ymin=0.6, ymax=1.0,
    ylabel={Hit@10},
    symbolic x coords={
        OMIM--ORDO,
        NCIT--DOID,
        SNOMED--NCIT (pharm),
        SNOMED--NCIT (neoplas),
        SNOMED--FMA (body)
    },
    xtick=data,
    x tick label style={rotate=35, anchor=east},
]
\addplot coordinates {
    (OMIM--ORDO, 0.909)
    (NCIT--DOID, 0.985)
    (SNOMED--NCIT (pharm), 0.966)
    (SNOMED--NCIT (neoplas), 0.932)
    (SNOMED--FMA (body), 0.947)
};
\addplot coordinates {
    (OMIM--ORDO, 0.907)
    (NCIT--DOID, 0.962)
    (SNOMED--NCIT (pharm), 0.946)
    (SNOMED--NCIT (neoplas), 0.938)
    (SNOMED--FMA (body), 0.930)
};
\end{axis}
\end{tikzpicture}
\caption{Hit@10}
\end{subfigure}

\caption{Impact of concept definitions on embedding-based candidate retrieval (\texttt{text-embedding-3-small}).}
\label{fig:embedding-definition-impact}
\end{figure}

\subsection{Effectiveness Compared to Original Exact Matching Methods}

We compared the results of GenOM (Qwen32B) with the results of the stand-alone exact matching system that GenOM (Qwen32B) adopted in the final stage (either BERTMapLt or LogMapLt). The BERTMapLt and LogMapLt results reported in this section were reproduced within the scope of this work and therefore might differ slightly from the results presented in Table~\ref{tab:bio_alignment_comparison}.

As shown in Table~\ref{tab:exactmatch_results}, GenOM (Qwen32B) surpassed the strongest exact-matching systems on most tasks, with the sole exception of the SNOMED--FMA (Body) alignment. In our earlier workshop version of this work~\cite{song2025genom}, which focused on 7B-level LLMs, we were able to outperform both BERTMapLt and LogMapLt across all tasks by imposing a relatively high cosine-similarity threshold. However, as discussed previously, larger LLMs exhibited much lower sensitivity to the cosine-similarity threshold, with Qwen-32B showing only minor fluctuations in F1 scores between thresholds of 0.6 and 0.9. Under such circumstances, enforcing extremely strict thresholds (e.g.~0.97), while effective for smaller models, is less suitable for larger ones, making a more balanced configuration preferable.

Even with this trade-off, GenOM (Qwen32B) consistently improved upon exact matching across the Bio-ML tasks. On average, GenOM (Qwen32B) achieved a \textbf{5.5\%} relative F1 improvement over 
\textbf{BERTMapLt} and a more substantial \textbf{15.7\%} improvement over \textbf{LogMapLt}. These gains demonstrated that the proposed framework remained effective even without relying on aggressive threshold filtering, and that the enhanced reasoning capabilities of larger LLMs could be translated into measurable performance improvements over strong symbolic baselines.

\begin{table}[ht]
\centering
\footnotesize
\setlength{\tabcolsep}{4pt}
\renewcommand{\arraystretch}{1}
\begin{tabular}{ll|ccc}
\toprule
\textbf{Task} & \textbf{Model Variant} &   \textbf{P}  &\textbf{R}  &\textbf{F1} \\
\midrule
\multirow{4}{*}{SNOMED-NCIT-neoplas} 
& GenOM(BERTMapLt)       & 0.747  & 0.804 & \textbf{0.774} \\
& BERTMapLt              & 0.831 & 0.687 & 0.752 \\
& GenOM(LogMapLt)       & 0.708 & 0.657 &  \textbf{0.681}  \\
& LogMapLt               & 0.952 & 0.491 & 0.648 \\
\midrule

\multirow{4}{*}{SNOMED-NCIT-pharm} 
& GenOM(BERTMapLt)       & 0.957 & 0.622 &  \textbf{0.754} \\
& BERTMapLt              & 0.981 & 0.574 & 0.724 \\
& GenOM(LogMapLt)       & 0.971 & 0.621  & \textbf{0.758} \\
& LogMapLt               & 0.996 & 0.586 & 0.738 \\   
\midrule

\multirow{4}{*}{SNOMED-FMA-body} 
& GenOM(BERTMapLt)        & 0.902 & 0.674 &  0.771 \\
& BERTMapLt                & 0.979 & 0.655 & \textbf{0.785} \\
& GenOM(LogMapLt)          & 0.766  & 0.557 & 0.645 \\
& LogMapLt                   & 0.971 & 0.527 & \textbf{0.683} \\   
\midrule

\multirow{4}{*}{OMIM-ORDO} 
& GenOM(BERTMapLt)          & 0.834 & 0.643 & \textbf{0.726} \\
& BERTMapLt                   & 0.834 & 0.497 & 0.623 \\
& GenOM(LogMapLt)            & 0.886  & 0.544 & \textbf{0.674} \\
& LogMapLt                     & 0.937 & 0.215 & 0.350\\ 
\midrule

\multirow{4}{*}{NCIT-DOID} 
& GenOM(BERTMapLt)           & 0.866 & 0.937 &  \textbf{0.901} \\
& BERTMapLt                 & 0.919 & 0.772 & 0.839 \\
& GenOM(LogMapLt)         & 0.914  & 0.876 & \textbf{0.895} \\
& LogMapLt                  & 0.955 & 0.602 & 0.738 \\ 
\bottomrule

\end{tabular}
\caption{The results of GenOM with Qwen32B using the exact matching systems BERTMapLt and LogMapLt, as well as the results of BERTMapLt and LogMapLt alone.}
\label{tab:exactmatch_results}
\end{table}



\section{Conclusion, Discussion and Future Work}\label{sec8}

This paper presents \textbf{GenOM}, a general-purpose framework for ontology alignment that integrated concept semantic enrichment with LLM-based textual definition generation, embedding-based candidate retrieval, LLM prompting-based equivalence judgement, and exact matching in a modular design. Evaluation results have shown the approach demonstrates strong performance across five biomedical ontology alignment tasks of OAEI Bio-ML, outperforming many baselines, and the effectiveness of its important components has been verified via ablation studies.
In particular, the framework shows its ability to generalise across datasets while maintaining alignment accuracy, without relying on handcrafted features or extensive task-specific engineering.

Although GemOM has achieved promising performance for equivalence mappings in OM, several challenges remain:
\begin{enumerate}

\item It is difficult to consistently assess the degree of equivalence between concept pairs. This challenge affects both the LLM-based judgement stage and the choice of cosine similarity threshold for candidate retrieval. 

\ignore{
\item The definition of equivalence can vary subtly across tasks: concept pairs deemed equivalent in one alignment task may not be considered so in another, leading to inconsistencies in judgement. 
}

\item The alignment performance of LLMs is highly sensitive to the prompt. In the evaluation, we observed that vague prompts such as simply asking the model to ``determine whether two concepts are equivalent'' often fails to elicit correct predictions; in many cases, the LLM almost never produces`'YES' output. This highlights the importance of prompt specificity in steering LLM behaviour and underscores a practical challenge in applying LLMs to alignment in a generalisable way.
\end{enumerate}

For future work, an important enhancement would be to expand the scope of GenOM to include additional alignment types beyond equivalence, such as subsumption. Another key direction would involves addressing the variability in how equivalence is defined across different ontologies and tasks. In many alignment scenarios, the threshold for considering two concepts equivalent may depend on contextual or domain-specific nuances, which are difficult to capture using a fixed similarity score or binary decision.
To tackle this, task-adaptive alignment criteria would be useful to explore, including dynamic threshold selection and prompt-based calibration techniques that allow the LLM to assess the strength or type of correspondence more flexibly. Additionally, incorporating finer-grained semantic similarity measures and confidence estimation strategies could help better reflect the spectrum of equivalence relations observed in practice.

\rev{Beyond ontology alignment, the proposed framework can potentially be extended to related semantic matching tasks, such as entity linking, where textual mentions are associated with ontology concepts. While the current work is specifically designed and evaluated for ontology alignment, the combination of definition generation and LLM-based reasoning could, in principle, be adapted to support mention-level semantic matching. We leave this as an interesting direction for future research.}

\clearpage
\section*{Declarations}

\textbf{Funding.} 
This research received no specific grant from any funding agency in the public, commercial, or not-for-profit sectors.
\\
\textbf{Data availability.} 
All datasets used in this study are publicly available from open-source ontology repositories.
\\
\textbf{Conflict of interest.} 
The authors declare that they have no competing interests.
\\
\textbf{Ethics approval.} 
This study does not involve human participants, animals, or any private or sensitive data and therefore does not require ethical approval.

\clearpage
\begin{appendices}

\section{Additional Experimental Details}\label{secA}

\subsection{Optimal Performance of GenOM(Qwen32B)}\label{secA1}
Table~\ref{tab:genom-upperbound} reports the per-task optimal performance that GenOM achieves when the decision thresholds are tuned separately for each ontology matching task. As expected, task-specific tuning leads to further improvements over the unified threshold setting reported in the main paper, reflecting the heterogeneous similarity distributions across different Bio-ML tracks. The results are not used for comparison in the study and do not replace the unified-threshold setting reported in the main text. They are provided as auxiliary evidence about the potential performance of GenOM(Qwen32B) under task-specific calibration when a set of uniformly sampled mappings are annotated for validation and hyper parameter searching.

\begin{table}[!ht]
\centering
\caption{Per-task optimal performance of GenOM(Qwen32B) under task-specific thresholds (Precision, Recall, and F1-score).}
\label{tab:genom-upperbound}
\begin{tabular}{lccc}
\toprule
\textbf{Bio-ML Track} & \textbf{Precision} & \textbf{Recall} & \textbf{F1-score} \\
\midrule
SNOMED-NCIT (Neoplas) &  0.812 & 0.765 & 0.788 \\
SNOMED-NCIT (Pharm) & 0.915 & 0.653 & 0.762 \\
SNOMED-FMA (Body) & 0.921 & 0.699 & 0.795 \\
OMIM-ORDO & 0.824 & 0.705 & 0.760 \\
NCIT-DOID & 0.873 & 0.938 & 0.904 \\
\bottomrule
\end{tabular}
\end{table}

\subsection{\rev{Detailed Ablation Study Results}}\label{secA2}

\begin{table}[ht]
\caption{Detailed Results of Definition-based Local Ranking (Qwen32B)}
\label{tab:localranking-definition-impact}
\centering
\footnotesize
\begin{tabular}{|l|c|c|}

\hline
\textbf{Task} & \textbf{MRR} & \textbf{Hit@1}  \\
\hline
OMIM\_ORDO (with) & 0.903  & 0.860 \\
OMIM\_ORDO (without) & 0.723 & 0.625 \\
NCIT\_DOID (with) & 0.949  & 0.924 \\
NCIT\_DOID (without) & 0.919  & 0.882 \\
SNOMED\_NCIT\_pharm (with) & 0.918  & 0.869 \\
SNOMED\_NCIT\_pharm (without) & 0.806  & 0.739 \\
SNOMED\_NCIT\_neoplas (with) & 0.928  & 0.888 \\
SNOMED\_NCIT\_neoplas (without) & 0.891 & 0.841 \\
SNOMED\_FMA\_body (with) & 0.871  & 0.819 \\
SNOMED\_FMA\_body (without) & 0.866  & 0.819 \\
\hline

\end{tabular}
\end{table}

\begin{table}[ht]
\centering
\footnotesize
\caption{Detailed Results of embedding-based candidate retrieval (text-embedding-3-small)}
\label{tab:embedding-definition-impact}
\begin{tabular}{|l|c|c|c|}
\hline
\textbf{Task} & \textbf{Hit@1} & \textbf{Hit@5} & \textbf{Hit@10} \\
\hline
OMIM\_ORDO (with) & 0.737 & 0.879 & 0.909 \\
OMIM\_ORDO (without) & 0.724 & 0.875 & 0.907 \\
NCIT\_DOID (with) & 0.886 & 0.974 & 0.985 \\
NCIT\_DOID (without) & 0.786 & 0.940 & 0.962 \\
SNOMED\_NCIT\_pharm (with) & 0.747 & 0.942 & 0.966 \\
SNOMED\_NCIT\_pharm (without) & 0.712 & 0.911 & 0.946 \\
SNOMED\_NCIT\_neoplas (with) & 0.722 & 0.896 & 0.932 \\
SNOMED\_NCIT\_neoplas (without) & 0.718 & 0.896 & 0.938 \\
SNOMED\_FMA\_body (with) & 0.695 & 0.912 & 0.947 \\
SNOMED\_FMA\_body (without) & 0.642 & 0.883 & 0.930 \\
\hline
\end{tabular}
\end{table}

\section{}\label{secB}
\subsection{Qualitative Sanity Check via Manual Inspection}\label{secB:manual_check}
As a qualitative sanity check, we conducted a follow-up manual inspection of the LLM-based evaluation outputs. 
Due to practical resource constraints, this inspection was limited to two datasets, Pharm-SNOMED and Body-FMA, covering a total of 200 biomedical concepts. 
The inspection was carried out by the authors using external resources, including publicly available biomedical references and search engines, and was further supplemented by consultation with general-purpose LLMs to assist in understanding unfamiliar medical concepts. 
Although the authors are not domain experts and did not examine all concepts evaluated by the LLM-based scheme, we observed a high degree of consistency between the manual assessments and the score produced by the automated evaluation within the inspected subset. 
This analysis was not intended as a formal validation, but rather as an auxiliary plausibility check to assess whether the proposed evaluation method produced rankings that were manifestly counter-intuitive.

\subsection{Prompt Template and Analysis}\label{secB:prompt_template}

During experimentation, we observed that the LLM occasionally interchanged the numerical values assigned to \texttt{discriminative} and \texttt{correctness} in its JSON output. This issue reflects a known limitation of LLMs in adhering to strict output schemas, 
particularly when multiple related scoring dimensions are requested simultaneously. In the present study, this behaviour does not affect the final evaluation results, because the two scores contribute equally (1:1 weighting) to the aggregated metric; swapping their positions leaves the combined score unchanged.  However, such inconsistencies would become consequential under non-uniform weighting schemes or alternative aggregation strategies.

\begin{table}[htbp]
\centering
\caption{Prompt used for LLM-based scoring of definition correctness and discriminativeness.}
\begin{tabular}{p{3cm} p{9cm}}
\toprule
\textbf{Section} & \textbf{Prompt Content} \\
\midrule

\textbf{Instruction} &
\verb|You are a biomedical ontology editor. Score four candidate | \\
&\verb|definitions for Concept A ONLY on:| \\
& \verb|- discriminative (1-5): helps distinguish A from concept B.| \\
& \verb|- correctness (1-5): medically plausible and not misleading.| \\[4pt]

\textbf{Output format} &
\verb|Return ONLY this JSON:| \\
& \verb|{{| \\
& \verb|  "scores": [| \\
& \verb|    {{"discriminative": int, "correctness": int}},| \\
& \verb|    {{"discriminative": int, "correctness": int}},| \\
& \verb|    {{"discriminative": int, "correctness": int}},| \\
& \verb|    {{"discriminative": int, "correctness": int}}| \\
& \verb|  ]| \\
& \verb|}}| \\[4pt]

\textbf{Concept A} &
\verb|## Concept A| \\
& \verb|label: "{a_label}"| \\
& \verb|definitions:| \\
& \verb|1) "{d1}"| \\
& \verb|2) "{d2}"| \\
& \verb|3) "{d3}"| \\
& \verb|4) "{d4}"| \\[4pt]

\textbf{Concept B} &
\verb|## Concept B| \\
& \verb|label: "{b_label}"| \\

\bottomrule
\end{tabular}
\end{table}

\section{\rev{Error Analysis of LLM-based Alignment}}\label{secC}

\rev{In this appendix, we present several representative error cases observed in the LLM-based equivalence judgement stage. The analysis focuses on two common error types: false positives, where the model incorrectly predicts equivalence between semantically related but non-equivalent concepts, and false negatives, where the model fails to identify correct equivalence due to differences in conceptual granularity or ontology modelling choices.}

\rev{It is also important to note that this OM task allows one-to-many and many-to-one mappings. As a result, the result is not restricted to selecting a single target concept for each source concept. This flexibility increases the likelihood of false positive predictions, since multiple semantically related concepts may be identified as potential matches.}

\clearpage
\subsection{False Positive Case}

\begin{table}[h]
\centering
\small
\begin{tabular}{p{3.5cm} p{9cm}}
\hline
\textbf{Source concept} & Biperiden hydrochloride \\
\textbf{Target concept} & Biperiden \\
\hline
\textbf{Observation} & Both concepts share highly similar semantic descriptions. The definitions emphasise that both substances act as anticholinergic agents used to treat symptoms associated with Parkinson's disease. Their parent classes are also closely related, including 'Antiparkinsonian agent' and 'Muscarinic cholinergic antagonist'. \\
\hline
\textbf{Error explanation} & The inclusion of detailed definitions further strengthens the perceived semantic similarity between the two concepts. As a result, the LLM incorrectly predicts equivalence. However, 'Biperiden hydrochloride' refers to a specific salt form of the drug, whereas 'Biperiden' denotes the base compound. Although pharmacologically related, they are modelled as distinct entities in the ontology and should therefore not be aligned as equivalent. \\
\hline
\textbf{Implication} & This example illustrates a potential side effect of definition-based semantic enrichment. Additional semantic information may amplify similarities between closely related biomedical concepts, increasing the likelihood of false positive alignments. \\
\hline
\end{tabular}
\caption{Representative false positive case in LLM-based ontology alignment.}
\end{table}

\clearpage
\subsection{False Negative Case}

\begin{table}[h]
\centering
\small
\begin{tabular}{p{3.5cm} p{9cm}}

\hline
\textbf{Source concept} & Quetiapine-containing product \\
\textbf{Target concept} & QUETIAPINE \\
\hline
\textbf{Observation} & The source concept refers to a medicinal product category that contains quetiapine as an active ingredient, whereas the target concept denotes the active pharmaceutical ingredient itself. The definitions and hierarchical information highlight that the two concepts belong to different levels of abstraction within the ontology. \\
\hline
\textbf{Error explanation} & From a semantic perspective, the model’s behaviour is understandable. One concept represents a class of medicinal products containing a particular ingredient, while the other represents the chemical substance itself. As a result, the LLM tends to interpret them as related but not strictly equivalent concepts. \\
\hline
\textbf{Implication} & According to the ground truth alignment provided in the benchmark dataset, these two concepts are considered equivalent. Therefore, the model prediction is counted as a false negative under the evaluation setting. This example suggests that some evaluation errors may arise from differences in conceptual granularity or ontology modelling choices rather than genuine semantic misunderstandings by the model. \\
\hline
\end{tabular}
\caption{Representative false negative case in LLM-based ontology alignment.}
\end{table}




\end{appendices}


\clearpage

\bibliography{sn-bibliography}

@String{Computing = "Computing" }

@String{Computer = "{IEEE} Computer" }

@String{Springer = "Springer-Verlag" }

@book{euzenat2013d,
        author          =       {J\'er\^ome Euzenat and Pavel Shvaiko},
        title           =       {Ontology matching},
        edition         =       {2nd},
        language        =       {english},
        page            =       520,
        publisher       =       {Springer-Verlag},
        address         =       {Heidelberg (DE)},
        year            =       2013}

@ARTICLE{6104044,
  author={Shvaiko, Pavel and Euzenat, Jérôme},
  journal={IEEE Transactions on Knowledge and Data Engineering}, 
  title={Ontology Matching: State of the Art and Future Challenges}, 
  year={2013},
  volume={25},
  number={1},
  pages={158-176},
  doi={10.1109/TKDE.2011.253}}

@misc{snomedct2024,
  author       = {{SNOMED International}},
  title        = {{SNOMED CT: The Global Clinical Terminology}},
  year         = {2024},
  howpublished = {\url{https://www.snomed.org}},
  note         = {Accessed: 2025-06-23}
}

@inproceedings{jimenez2011logmap,
  title={Logmap: Logic-based and scalable ontology matching},
  author={Jim{\'e}nez-Ruiz, Ernesto and Cuenca Grau, Bernardo},
  booktitle={The Semantic Web--ISWC 2011: 10th International Semantic Web Conference, Bonn, Germany, October 23-27, 2011, Proceedings, Part I 10},
  pages={273--288},
  year={2011},
  organization={Springer}
}

@inproceedings{kolyvakis-etal-2018-deepalignment,
    title = "{D}eep{A}lignment: Unsupervised Ontology Matching with Refined Word Vectors",
    author = "Kolyvakis, Prodromos  and
      Kalousis, Alexandros  and
      Kiritsis, Dimitris",
    editor = "Walker, Marilyn  and
      Ji, Heng  and
      Stent, Amanda",
    booktitle = "Proceedings of the 2018 Conference of the North {A}merican Chapter of the Association for Computational Linguistics: Human Language Technologies, Volume 1 (Long Papers)",
    month = jun,
    year = "2018",
    address = "New Orleans, Louisiana",
    publisher = "Association for Computational Linguistics",
    url = "https://aclanthology.org/N18-1072/",
    doi = "10.18653/v1/N18-1072",
    pages = "787--798"
}

@inproceedings{faria2013agreementmakerlight,
  title={The agreementmakerlight ontology matching system},
  author={Faria, Daniel and Pesquita, Catia and Santos, Emanuel and Palmonari, Matteo and Cruz, Isabel F and Couto, Francisco M},
  booktitle={On the Move to Meaningful Internet Systems: OTM 2013 Conferences: Confederated International Conferences: CoopIS, DOA-Trusted Cloud, and ODBASE 2013, Graz, Austria, September 9-13, 2013. Proceedings},
  pages={527--541},
  year={2013},
  organization={Springer}
}

@inproceedings{bento2020ontology,
  title={Ontology matching using convolutional neural networks},
  author={Bento, Alexandre and Zouaq, Amal and Gagnon, Michel},
  booktitle={Proceedings of the Twelfth Language Resources and Evaluation Conference},
  pages={5648--5653},
  year={2020}
}

@inproceedings{he2022bertmap,
  title="{BERTM}ap: a {BERT}-based ontology alignment system",
  author={He, Yuan and Chen, Jiaoyan and Antonyrajah, Denvar and Horrocks, Ian},
  booktitle={Proceedings of the AAAI Conference on Artificial Intelligence},
  volume={36},
  number={5},
  pages={5684--5691},
  year={2022}
}

@inproceedings{devlin-etal-2019-bert,
    title = "{BERT}: Pre-training of Deep Bidirectional Transformers for Language Understanding",
    author = "Devlin, Jacob  and
      Chang, Ming-Wei  and
      Lee, Kenton  and
      Toutanova, Kristina",
    editor = "Burstein, Jill  and
      Doran, Christy  and
      Solorio, Thamar",
    booktitle = "Proceedings of the 2019 Conference of the North {A}merican Chapter of the Association for Computational Linguistics: Human Language Technologies, Volume 1 (Long and Short Papers)",
    month = jun,
    year = "2019",
    address = "Minneapolis, Minnesota",
    publisher = "Association for Computational Linguistics",
    url = "https://aclanthology.org/N19-1423/",
    doi = "10.18653/v1/N19-1423",
    pages = "4171--4186"
}

@article{Norouzi2023ConversationalOA,
  title={Conversational ontology alignment with chatgpt},
  author={Norouzi, Sanaz Saki and Mahdavinejad, Mohammad Saeid and Hitzler, Pascal},
  journal={arXiv preprint arXiv:2308.09217},
  year={2023}
}

@article{giglou2024llms4om,
  title={Llms4om: Matching ontologies with large language models},
  author={Giglou, Hamed Babaei and D'Souza, Jennifer and Engel, Felix and Auer, S{\"o}ren},
  journal={arXiv preprint arXiv:2404.10317},
  year={2024}
}

@inproceedings{he2022machine,
  title={Machine learning-friendly biomedical datasets for equivalence and subsumption ontology matching},
  author={He, Yuan and Chen, Jiaoyan and Dong, Hang and Jim{\'e}nez-Ruiz, Ernesto and Hadian, Ali and Horrocks, Ian},
  booktitle={International Semantic Web Conference},
  pages={575--591},
  year={2022},
  organization={Springer}
}

@inproceedings{hertling2023olala,
  title={Olala: Ontology matching with large language models},
  author={Hertling, Sven and Paulheim, Heiko},
  booktitle={Proceedings of the 12th Knowledge Capture Conference 2023},
  pages={131--139},
  year={2023}
}

@article{he2023exploring,
  title={Exploring large language models for ontology alignment},
  author={He, Yuan and Chen, Jiaoyan and Dong, Hang and Horrocks, Ian},
  journal={arXiv preprint arXiv:2309.07172},
  year={2023}
}

@article{10.14778/3712221.3712222,
author = {Qiang, Zhangcheng and Wang, Weiqing and Taylor, Kerry},
title = "{A}gent-{OM}: Leveraging {LLM} Agents for Ontology Matching",
year = {2024},
issue_date = {November 2024},
publisher = {VLDB Endowment},
volume = {18},
number = {3},
issn = {2150-8097},
url = {https://doi.org/10.14778/3712221.3712222},
doi = {10.14778/3712221.3712222},
journal = {Proc. VLDB Endow.},
month = nov,
pages = {516–529},
numpages = {14}
}

@article{DBLP:journals/semweb/HeCDHAKS24,
  author = {Yuan He and Jiaoyan Chen and Hang Dong and Ian Horrocks
		  and Carlo Allocca and Taehun Kim and Brahmananda Sapkota},
  bibsource = {dblp computer science bibliography, https://dblp.org},
  doi = {10.3233/SW-243568},
  journal = {Semantic Web},
  number = {5},
  pages = {1991--2004},
  title = {{DeepOnto}: {A} Python package for ontology engineering
		  with deep learning},
  url = {download/2024/HeCDHAKS24.pdf},
  volume = {15},
  year = {2024}
}

@article{DBLP:journals/www/ChenHGJDH23,
  author       = {Jiaoyan Chen and
                  Yuan He and
                  Yuxia Geng and
                  Ernesto Jim{\'{e}}nez{-}Ruiz and
                  Hang Dong and
                  Ian Horrocks},
  title        = {Contextual semantic embeddings for ontology subsumption prediction},
  journal      = {World Wide Web {(WWW)}},
  volume       = {26},
  number       = {5},
  pages        = {2569--2591},
  year         = {2023},
  url          = {https://doi.org/10.1007/s11280-023-01169-9},
  doi          = {10.1007/S11280-023-01169-9},
  bibsource    = {dblp computer science bibliography, https://dblp.org}
}

@article{anam2015review,
  title={Review of ontology matching approaches and challenges},
  author={Anam, Sarawat and Kim, Yang Sok and Kang, Byeong Ho and Liu, Qing},
  journal={International Journal of Computer Science and Network Solutions},
  volume={3},
  number={3},
  pages={1--27},
  year={2015}
}

@book{staab2013handbook,
  title        = {Handbook on Ontologies},
  author       = {Staab, Steffen and Studer, Rudi},
  year         = {2013},
  publisher    = {Springer},
  address      = {Berlin, Heidelberg}
}

@inproceedings{nkisi2018ontology,
  title={Ontology alignment based on word embedding and random forest classification},
  author={Nkisi-Orji, Ikechukwu and Wiratunga, Nirmalie and Massie, Stewart and Hui, Kit-Ying and Heaven, Rachel},
  booktitle={Joint European Conference on Machine Learning and Knowledge Discovery in Databases},
  pages={557--572},
  year={2018},
  organization={Springer}
}

@incollection{doan2004ontology,
  author    = {Doan, Anhai and Madhavan, Jayant and Domingos, Pedro and Halevy, Alon},
  title     = {Ontology Matching: A Machine Learning Approach},
  booktitle = {Handbook on Ontologies},
  editor    = {Staab, Steffen and Studer, Rudi},
  publisher = {Springer},
  address   = {Berlin, Heidelberg},
  pages     = {385--403},
  year      = {2004},
  doi       = {10.1007/978-3-540-24750-0_19}
}

@inproceedings{wang2018ontology,
  title={Ontology alignment in the biomedical domain using entity definitions and context},
  author={Wang, Lucy Lu and Bhagavatula, Chandra and Neumann, Mark and Lo, Kyle and Wilhelm, Chris and Ammar, Waleed},
  booktitle={Proceedings of the BioNLP 2018 workshop},
  pages={47--55},
  year={2018}
}

@inproceedings{chen2021augmenting,
  title={Augmenting ontology alignment by semantic embedding and distant supervision},
  author={Chen, Jiaoyan and Jim{\'e}nez-Ruiz, Ernesto and Horrocks, Ian and Antonyrajah, Denvar and Hadian, Ali and Lee, Jaehun},
  booktitle={European Semantic Web Conference},
  pages={392--408},
  year={2021},
  organization={Springer}
}

@inproceedings{menad2023biostransformers,
  title     = "{B}io{ST}ransformers for Biomedical Ontologies Alignment",
  author    = {Menad, S. and Laddada, W. and Abdeddaim, S. and Soualmia, L. F.},
  booktitle = {Proceedings of the 15th International Joint Conference on Knowledge Discovery, Knowledge Engineering and Knowledge Management (IC3K 2023)},
  pages     = {73--84},
  year      = {2023},
  doi       = {10.5220/0012188600003598}
}

@article{reimers2019sentence,
  title="Sentence-{BERT}: Sentence embeddings using {S}iamese {BERT}-networks",
  author={Reimers, Nils and Gurevych, Iryna},
  journal={arXiv preprint arXiv:1908.10084},
  year={2019}
}

@inproceedings{song2025genom,
  title={GenOM: Ontology Matching with Description Generation and Large Language Model},
  author={Song, Yiping and Chen, Jiaoyan and Schmidt, Renate A},
  booktitle = {Proceedings of the Ontology Matching Workshop (OM) at the 24th International Semantic Web Conference (ISWC 2025)},
  year      = {2025}
}

@article{lushnei2025large,
  title={Large Language Models as Oracles for Ontology Alignment},
  author={Lushnei, Sviatoslav and Shumskyi, Dmytro and Shykula, Severyn and Jimenez-Ruiz, Ernesto and Garcez, Artur d'Avila},
  journal={arXiv preprint arXiv:2508.08500},
  year={2025}
}

@inproceedings{hu2025matching,
  author    = {Hu, Wenxin and Ichise, Ryutaro},
  title     = "From Matching to Retrieval: A New Role for {LLM}s in Ontology Alignment",
  booktitle = {Proceedings of the Ontology Matching Workshop (OM) at the 24th International Semantic Web Conference (ISWC 2025)},
  year      = {2025}
}

@inproceedings{liu2021self,
  title={Self-alignment pretraining for biomedical entity representations},
  author={Liu, Fangyu and Shareghi, Ehsan and Meng, Zaiqiao and Basaldella, Marco and Collier, Nigel},
  booktitle={Proceedings of the 2021 conference of the North American chapter of the association for computational linguistics: human language technologies},
  pages={4228--4238},
  year={2021}
}

@article{johnson2019billion,
  title="Billion-scale similarity search with {GPUs}",
  author={Johnson, Jeff and Douze, Matthijs and J{\'e}gou, Herv{\'e}},
  journal={IEEE Transactions on Big Data},
  volume={7},
  number={3},
  pages={535--547},
  year={2019},
  publisher={IEEE}
}

@article{kadavath2022language, 
    title={Language models (mostly) know what they know}, author={Kadavath, Saurav and Conerly, Tom and Askell, Amanda and Henighan, Tom and Drain, Dawn and Perez, Ethan and Schiefer, Nicholas and Hatfield-Dodds, Zac and DasSarma, Nova and Tran-Johnson, Eli and others}, 
    journal={arXiv preprint arXiv:2207.05221}, 
    year={2022} 
}

@article{si2022prompting, 
    title={Prompting gpt-3 to be reliable}, 
    author={Si, Chenglei and Gan, Zhe and Yang, Zhengyuan and Wang, Shuohang and Wang, Jianfeng and Boyd-Graber, Jordan and Wang, Lijuan}, 
    journal={arXiv preprint arXiv:2210.09150}, 
    year={2022} 
}

@inproceedings{guo2017calibration, 
    title={On calibration of modern neural networks}, author={Guo, Chuan and Pleiss, Geoff and Sun, Yu and Weinberger, Kilian Q}, 
    booktitle={International conference on machine learning}, 
    pages={1321--1330}, 
    year={2017}, 
    organization={PMLR} 
}

@inproceedings{jimenez2015logmap,
  title="{LogMap} family results for {OAEI} 2015.",
  author={Jim{\'e}nez-Ruiz, Ernesto and Grau, Bernardo Cuenca and Solimando, Alessandro and Cross, Valerie V},
  booktitle={OM},
  pages={171--175},
  year={2015}
}

@inproceedings{10.1145/3746252.3761533,
author = {Luo, Weiqing and Song, Chonggang and Yi, Lingling and Cheng, Gong},
title = {TRAWL: External Knowledge-Enhanced Recommendation with LLM Assistance},
year = {2025},
isbn = {9798400720406},
publisher = {Association for Computing Machinery},
address = {New York, NY, USA},
url = {https://doi.org/10.1145/3746252.3761533},
doi = {10.1145/3746252.3761533},
booktitle = {Proceedings of the 34th ACM International Conference on Information and Knowledge Management},
pages = {5907–5914},
numpages = {8},
location = {Seoul, Republic of Korea},
series = {CIKM '25}
}

@incollection{lambrix2009information,
  author    = {Lambrix, Patrick and Str{\"o}mb{\"a}ck, Lena and Tan, He},
  title     = {Information Integration in Bioinformatics with Ontologies and Standards},
  booktitle = {Semantic Techniques for the Web},
  editor    = {Bry, Fran{\c{c}}ois and Matuszy{\'n}ski, Jan},
  series    = {Lecture Notes in Computer Science},
  volume    = {5500},
  publisher = {Springer},
  address   = {Berlin, Heidelberg},
  year      = {2009},
  doi       = {10.1007/978-3-642-04581-3_8},
  isbn      = {978-3-642-04581-3}
}

@inproceedings{liu_are_2026,
	address = {Cham},
	title = {Are {LLMs} {Really} {Knowledgeable} for {Knowledge} {Graph} {Completion}?},
	isbn = {978-3-032-09530-5},
	booktitle = {The {Semantic} {Web} – {ISWC} 2025},
	publisher = {Springer Nature Switzerland},
	author = {Liu, Yang and Sun, Zequn and Shao, Zhoutian and Cui, Yuanning and Hu, Wei},
	editor = {Garijo, Daniel and Kirrane, Sabrina and Salatino, Angelo and Shimizu, Cogan and Acosta, Maribel and Nuzzolese, Andrea Giovanni and Ferrada, Sebastián and Soulard, Thibaut and Kozaki, Kouji and Takeda, Hideaki and Gentile, Anna Lisa},
	year = {2026},
	pages = {77--95},
}

@incollection{delpinto2024ips,
  title     = {International Patient Summary Terminology},
  author    = {Del-Pinto, Warren and Schmidt, Renate A. and Gao, Yongsheng and Alghamdi, Ghadah and Lopez Osornio, Alejandro and Roy, Suzy},
  booktitle = {MEDINFO 2023 -- The Future Is Accessible},
  editor    = {Bichel-Findlay, J. and others},
  series    = {Studies in Health Technology and Informatics},
  volume    = {310},
  pages     = {63--67},
  year      = {2024},
  publisher = {IOS Press},
  address   = {Amsterdam},
  doi       = {10.3233/SHTI230928}
}

@inproceedings{del-pinto_extracting_2022,
	address = {Cham},
	title = {Extracting {Subontologies} from {SNOMED} {CT}},
	isbn = {978-3-031-11609-4},
	booktitle = {The {Semantic} {Web}: {ESWC} 2022 {Satellite} {Events}},
	publisher = {Springer International Publishing},
	author = {Del-Pinto, Warren and Schmidt, Renate A. and Gao, Yongsheng},
	editor = {Groth, Paul and Rula, Anisa and Schneider, Jodi and Tiddi, Ilaria and Simperl, Elena and Alexopoulos, Panos and Hoekstra, Rinke and Alam, Mehwish and Dimou, Anastasia and Tamper, Minna},
	year = {2022},
	pages = {291--294},
}

@article{brown2020language,
  title={Language models are few-shot learners},
  author={Brown, Tom and Mann, Benjamin and Ryder, Nick and Subbiah, Melanie and Kaplan, Jared D and Dhariwal, Prafulla and Neelakantan, Arvind and Shyam, Pranav and Sastry, Girish and Askell, Amanda and others},
  journal={Advances in neural information processing systems},
  volume={33},
  pages={1877--1901},
  year={2020}
}
\end{document}